\documentclass[twocolumn]{autart}
\input epsf
\usepackage{cite,epsfig}
\usepackage{graphicx,amsmath,amssymb,remark,color}
\usepackage[T1]{fontenc}
\usepackage[latin1]{inputenc}
\usepackage{graphicx}% Include figure files
\usepackage{amssymb}
\usepackage{amsmath}
\usepackage{dcolumn}% Align table columns on decimal point
\usepackage{bm}% bold math
\usepackage{color}
\usepackage{multicol}
\usepackage{algorithm}
\usepackage{multirow}
\usepackage{subfigure} 
\usepackage{hyperref}
\usepackage{xcolor}
\usepackage{xpatch}
\input epsf
\usepackage{cite,epsfig}
\usepackage{graphicx,amsmath,amssymb,remark,color}
\newcommand{\EQ}{\begin{eqnarray}}
\newcommand{\EN}{\end{eqnarray}}
\newcommand{\EQQ}{\begin{eqnarray*}}
\newcommand{\ENN}{\end{eqnarray*}}
\def\widebar{\accentset{{\cc@style\underline{\mskip10mu}}}}

 \makeatletter
\newcommand*\bigcdot{\mathpalette\bigcdot@{.5}}
\newcommand*\bigcdot@[2]{\mathbin{\vcenter{\hbox{\scalebox{#2}{$\m@th#1\bullet$}}}}}
\makeatother

\DeclareMathAlphabet{\matheur}{U}{eur}{m}{n}
\DeclareMathAlphabet{\matheurb}{U}{eur}{b}{n}
\DeclareMathAlphabet{\matheus}{U}{eus}{m}{n}
\DeclareMathAlphabet{\matheuf}{U}{euf}{m}{n}
\renewcommand{\t}{^{\mbox{\tiny\sf T}}}

\newcommand{\cal}{\mathcal}
\newcommand{\eproof}{\hfill\rule{2mm}{2mm}}

\newcommand{\bremark}{\begin{remark}
\begin{rm}}
\newcommand{\eremark}{ \end{rm}
\end{remark} }
\newcommand{\btheorem}{\begin{theorem} \begin{it}}
\newcommand{\etheorem}{\end{it} \hfill \rule{1mm}{2mm}
\end{theorem} }
\newcommand{\blemma}{\begin{lemma} \begin{it} }
\newcommand{\elemma}{ \end{it} \hfill\rule{1mm}{2mm}
\end{lemma} }
\newcommand{\bcorollary}{\begin{corollary} \begin{it} }
\newcommand{\ecorollary}{ \end{it} \hfill\rule{1mm}{2mm}
\end{corollary} }
\newcommand{\bdefinition}{\begin{definition} }
\newcommand{\edefinition}{ \hfill\rule{1mm}{2mm}
\end{definition} }
\newcommand{\bproposition}{\begin{proposition} }
\newcommand{\eproposition}{\hfill \rule{1mm}{2mm}
\end{proposition} }
\newcommand{\bexample}{\begin{example} \begin{rm}}
\newcommand{\eexample}{ \end{rm} \hfill\rule{1mm}{2mm}
\end{example} }

\newcommand{\basm}{\begin{assumption} \begin{rm} }
\newcommand{\easm}{ \end{rm} \hfill\rule{1mm}{2mm}
\end{assumption} }

\begin{document}

\newtheorem{theorem}{\sf\bfseries Theorem}[section]
\newtheorem{lemma}{\sf\bfseries Lemma}[section]
\newtheorem{coro}{\sf\bfseries Corollary}[section]
\newtheorem{definition}{\sf\bfseries Definition}[section]
\newtheorem{remark}{\sf\bfseries Remark}[section]
\newtheorem{corollary}{\sf\bfseries Corollary}[section]
\newtheorem{proposition}{\sf\bfseries Proposition}[section]
\newtheorem{example}{\sf\bfseries Example}[section]
\newtheorem{assumption}{\sf\bfseries Assumption}

\begin{frontmatter}

\title{  Ordering-Flexible Multi-Robot Coordination for \\Moving Target Convoying  Using Long-Term Task Execution }

\author[ntu]{Bin-Bin Hu},
\author[ntu]{Yanxin Zhou}, 
\author[ntu]{Henglai Wei},
\author[ntu]{Yan Wang}, 
\author[ntu]{Chen Lv}$^*$

\address[ntu]
{School of Mechanical and Aerospace Engineering\\
  Nanyang Technological University\\
 Singapore 637460\\
Email:  {\tt \{binbin.hu, yanxin.zhou, henglai.wei, yan\_wang, lyuchen\}@ntu.edu.sg}\\  
$^*$ Corresponding author}

\thanks{. This work was supported in part by the StartUp Grant-Nanyang Assistant Professorship Grant of Nanyang Technological University, in part by the Agency for Science, Technology and Research (A*STAR), Singapore, under Advanced Manufacturing and Engineering (AME) Young Individual Research under Grant A2084c0156, in
part by the MTC Individual Research under Grant M22K2c0079, in part
by the ANR-NRF Joint Grant NRF2021-NRF-ANR003 HM Science, and
in part by the Ministry of Education (MOE), Singapore, under the Tier 2
Grant MOE-T2EP50222-0002. Paper no. TII-23-3031. } 
 

\begin{abstract}
In this paper, we propose a cooperative long-term task execution (LTTE) algorithm for protecting a moving target into the interior of an {\it ordering-flexible} convex hull by a team of robots resiliently in the changing environments. Particularly, by designing target-approaching and sensing-neighbor collision-free subtasks, and incorporating these subtasks into the constraints rather than the traditional cost function in an online constraint-based optimization framework, the proposed LTTE can systematically guarantee {\it long-term} target convoying under {\it changing environments} in the $n$-dimensional Euclidean space. Then, the introduction of slack variables allow for the constraint violation of different subtasks; i.e., the attraction from target-approaching constraints and the repulsion from {\it time-varying} collision-avoidance constraints, which results in the desired formation with arbitrary spatial ordering sequences. Rigorous analysis is provided to guarantee asymptotical convergence with challenging nonlinear couplings induced by {\it time-varying} collision-free constraints. Finally, 2D experiments using three autonomous mobile robots (AMRs) are conducted to validate the effectiveness of the proposed algorithm, and 3D simulations tackling changing environmental elements, such as different initial positions, some robots suddenly breakdown and static obstacles are presented to demonstrate the multi-dimensional adaptability, robustness and the ability of obstacle avoidance of the proposed method.
\end{abstract}

\end{frontmatter}

\section{Introduction}

The deployment of a group of robots collectively providing protection for a target, called {\it multi-robot target convoying}, is playing an increasingly important role in complex urban missions, such as highway-vehicle conveyance \cite{han2022distributed},  marine vessel escort \cite{hu2021distributed1} and  multi-UAV protection \cite{islam2021robot}. In all these applications, the key component is to design a coordinated convoying protocol for robots to collectively form a convex hull such that the target can be eventually protected in the interior of the hull.

To form such a convex hull, predefining the desired robot-target relative positions (or displacements) with a fixed spatial ordering offers a straightforward and convenient approach, which is referred to as ordering-fixed target convoying. In this effort, early works focused on forming a fixed-ordering convoying formation for a static target \cite{chen2010surrounding}. Later, it was extended to a moving target \cite{hu2021bearing,shi2015cooperative}. For more complex multiple moving targets, a distributed convoying controller based on the estimation of targets' center was designed in \cite{hu2020multiple}. However, the aforementioned ordering-fixed convoying works \cite{chen2010surrounding,hu2021bearing,shi2015cooperative,hu2020multiple} stipulate the desired position and ordering of each robot in the controller setup in advance, which is not flexible in dynamic changing environments. It may waste more energy and lead to deadlock. For instance, in a $4$-robot target convoying task, if the initial position of robot~$1$ is far behind robots $2, 3, 4$ but the desired convoying position of robot $1$ is designed at the front of the hull, then robot~$1$ needs to go through all other robots  
to approach the desired position, which inevitably consumes more time and energy. Moreover, if the initial positions of robots $1, 2$ are on the both side of the target but the desired positions of robots $1, 2$ are on the opposite directions, it may cause the deadlock and failure of convoying.

In contrast, another research direction in the literature leverages a more efficient {\it ordering-flexible} approach to convoy the target into a convex hull with arbitrary spatial orderings~\cite{sakurama2020multi}, which
can mitigate the inflexibility and potential deadlock in the previous ordering-fixed setup \cite{chen2010surrounding,hu2021bearing,shi2015cooperative,hu2020multiple}. Essentially, the spatial orderings of the convex hull are not fixed and specified in advance, which are determined by the inter-robot interaction during the convoying process.
As the pioneering work, an output-regulation algorithm was proposed in \cite{liu2019collective} to convoy a static target. A limit-cycle-based decoupled structure was developed in \cite{wang2017limit} to convoy a static target with additional collision avoidance. Later, it was extended to a constant-velocity {\it ordering-flexible} target convoying \cite{hu2021distributed2,kou2021cooperative1,kou2021cooperative2}. For the variational-velocity target, a dynamic regulator with an internal model was proposed in \cite{hu2022cooperative} to maintain a rigid convoying formation. However, the previous works~\cite{sakurama2020multi,liu2019collective,wang2017limit,hu2021distributed2,kou2021cooperative1,kou2021cooperative2,hu2022cooperative} only do the largest effort to achieve the {\it ordering-flexible} target convoying vaguely, but still fail to provide a paradigm to guarantee whether {\it ordering-flexible} convoying problem is feasible or not, which thus 
may not work in the changing environments. Moreover, 
the previous {\it ordering-flexible} works \cite{sakurama2020multi,liu2019collective,wang2017limit,hu2021distributed2,kou2021cooperative1,kou2021cooperative2,hu2022cooperative} only study the target convoying tasks in open 2D environments, the exploration of higher-dimensional {\it ordering-flexible} target convoying in more complex environments still remains an open problem.

Motivated by the long-term task execution framework in~\cite{notomista2019constraint,notomista2021resilient}, we systematically bridge the gap between the feasibility of  {\it ordering-flexible} convoy and the online constraint-based optimization framework by proposing a LTTE algorithm for a multi-robot system to convoy a moving target into an {\it ordering-flexible} convex hull resiliently in the $n$-dimensional Euclidean space. More precisely, we design the target-approaching and sensing-neighbor collision-free subtasks, and encode such subtasks as constraints in the optimization framework for {\it long-term} target convoying under {\it changing environments.} 
The slack variables are then introduced to allow for the violation of different subtask constraints, namely, the attraction from target-approaching constraints and the repulsion from time-varying collision-avoidance constraints, which help form the convoying formation with arbitrary spatial ordering sequences. Finally, 2D experiments and 3D simulations are conducted to verify the effectiveness, multi-dimensional adaptability, robustness and the ability of obstacle avoidance of the proposed LTTE algorithm.

The main contribution of this paper is threefold.
\begin{enumerate}
 \item We propose a LTTE algorithm to establish an online constraint-based optimization framework for achieving multi-robot {\it long-term} target convoying with
an arbitrary ordering in changing environments.

 \begin{figure}[!htb]
\centering
\includegraphics[width=7cm]{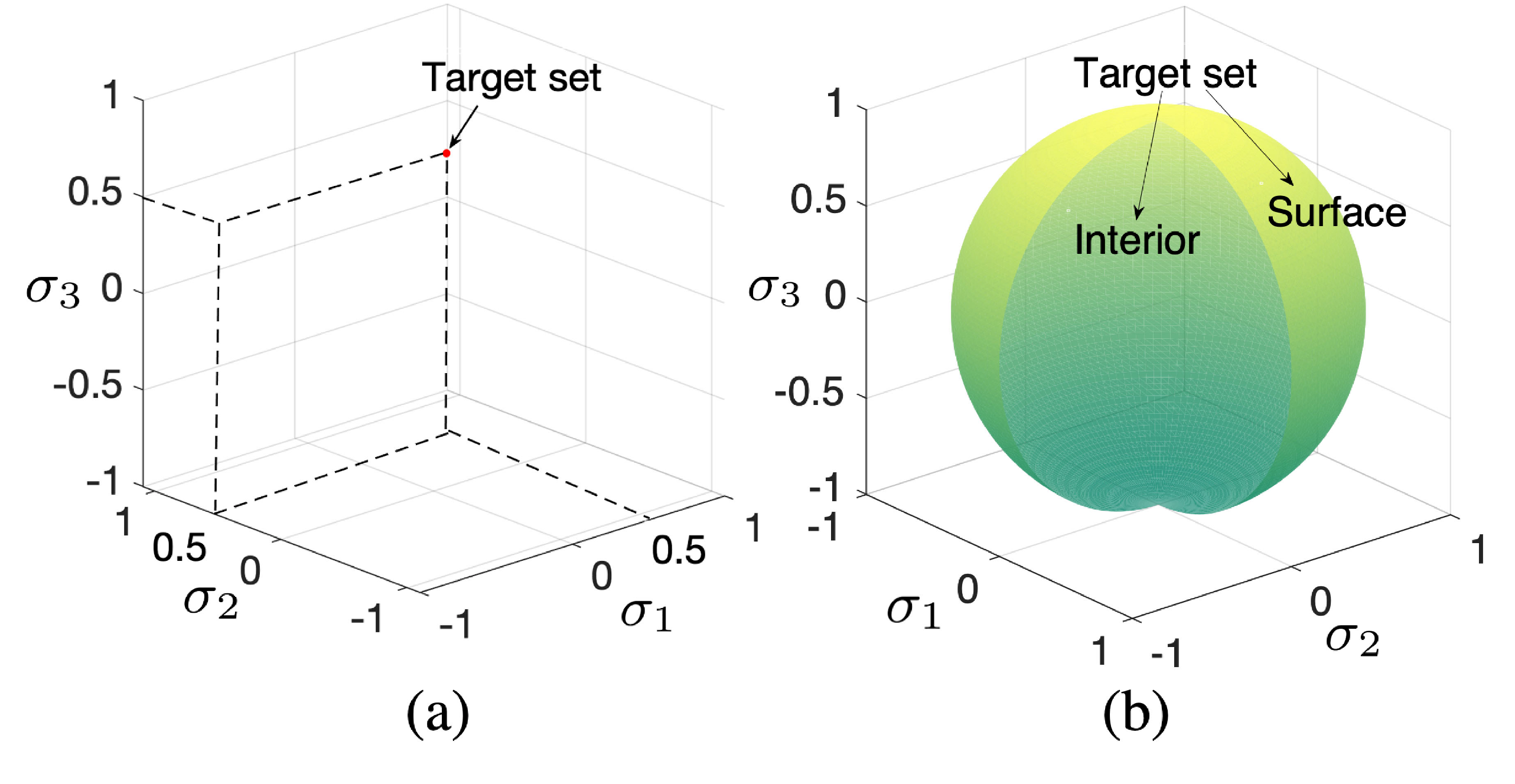}
\caption{(a) A point-target set $\mathcal T: \phi(\sigma):=(\sigma_1-0.5)^2$ $+(\sigma_2-0.5)^2+(\sigma_3-0.5)^2\leq0$ only contains a red point $p_0=[0.5, 0.5, 0.5]\t$. (b) A ball-target set $\mathcal T: \phi(\sigma)=\sigma_1^2$ $+\sigma_2^2+\sigma_3^2-1\leq0$ contains both the surface and interior of the green ball.
}
\label{Target_illustration}
\end{figure}
 
 \item We allow for the violation of different subtask constraints by introducing slack variables in the constraints, and rigorously guarantee the asymptotical convergence to {\it ordering-flexible} target convoying with strong nonlinear couplings induced by time-varying collision-free constraints.

\item We demonstrate the effectiveness of the proposed approach through 2D {\it ordering-flexible} target-convoying experiments of three AMRs, and the multi-dimensional adaptability, robustness and the ability of obstacle avoidance by 3D simulations tackling different initial positions, some robots suddenly breakdown and static obstacles.

\end{enumerate}

The remainder of this paper is organized as follows. In Section~2, we introduce the definitions and problem formulation concerning the long-term task execution and {\it ordering-flexible} target convoying, while the corresponding analysis of asymptotic convergence is presented in Section 3. 2D experiments and 3D simulations are shown in Section~4. Finally, conclusion is drawn in Section 5.

Throughout the paper, the real numbers and positive real numbers are denoted by $\mathbb{R},\mathbb{R}^+$, respectively. The $n$-dimensional Euclidean space is denoted by $\mathbb{R}^n$. The integer numbers are denoted by $\mathbb{Z}$. The notation $\mathbb{Z}_i^j$ represents the integer set $\{m\in \mathbb{Z}~|~i\leq m\leq j\}$. The $n$-dimensional identity matrix is represented by~$I_n$. The $n$-dimensional column vector consisting of all 0's is denoted by $\mathbf{0}_n$.

\section{Problem Formulation}
\subsection{Long-Term Task Execution}
Suppose a task set $\mathcal T$ is described by an implicit function $\phi(\sigma)$ \cite{yao2021singularity}
\begin{align}
\label{task_set}
\mathcal T :=& \{\sigma\in\mathbb{R}^d~|~\phi(\sigma)\leq0\},
\end{align}
where $\sigma=[\sigma_1, \cdots, \sigma_d]\t\in\mathbb{R}^d$ are the dummy coordinates with $d\in\mathbb{R}^+$ being its dimension, and $\phi(\cdot): \mathbb{R}^d\rightarrow\mathbb{R}$ is twice continuously differentiable (i.e., $\phi(\cdot)\in C^2$). $\phi(\sigma)\leq 0$ in Eq.~\eqref{task_set} is used to describe the region of the target set $\mathcal T$, which contain all possible states that belong to $\mathcal T$.

\begin{remark}
The formulation in Eq.~\eqref{task_set} can describe the target set $\mathcal T$ as a point or 1D(2D) manifold region in a topological manner with different selections of $\phi(\sigma)$. As shown in Fig.~\ref{Target_illustration}, a point-target set is defined to be $\phi(\sigma, p_0):=\|\sigma-p_0\|$ with $p_0=[0.5,0.5,0.5]\t$, and a target ball is set to be $\phi(\sigma):=\sigma_1^2+\sigma_2^2+\sigma_3^2-r^2$ with the radius $r=1$. Moreover, different from the sophisticated calculation of distance $\mbox{dist}(p_1, \mathcal T):=\inf \{\|p-p_1\| \big| p\in\mathcal T\}$ between a point $p_1\in\mathbb{R}^d$ and the target set $\mathcal T$, $\phi(p_1)$ in Eq.~\eqref{task_set} instead provides an implicit method to measure the distance to the target set $\mathcal T$ conveniently. An intuitive example is
the aforementioned target ball $\phi(\sigma):=\sigma_1^2+\sigma_2^2+\sigma_3^2-r^2$, where $\phi(p_1)>0$ if the point $p_1$ is at outside of the ball, otherwise $\phi(p_1)\leq0$ if the point $p_1$ is at the surface and interior of the ball. However, there exist some pathological situations which undermine such implicit measurements, i.e., $\lim_{t\rightarrow\infty}\phi(p_1(t))\leq0\nRightarrow\lim_{t\rightarrow\infty}\mbox{dist}(p_1(t), \mathcal T)=0$. The following assumption will be utilized to exclude these pathological situations.
\end{remark}

\begin{assumption}
\label{assump_exclusion}
\cite{yao2021singularity} For any given $\kappa>0$ and a point $p_1(t)\in\mathbb{R}^d$, one has that $\inf \{\phi(p_1(t)): \mathrm{dist}(p_1(t), \mathcal T)\geq\kappa\}>0.$
\end{assumption}

Assumption~\ref{assump_exclusion} ensures the implicit function $\phi(p_1(t))$ being utilized to rigorously represent the distance measurement from the point $p_1$ to the target set $\mathcal T$, which can be satisfied by some polynomial functions \cite{yao2021singularity}.

Let $x\in\mathbb{R}^d$ be the state vector of the robot. Substituting $x$ into the parameterization of $\mathcal T$ in Eq.~\eqref{task_set}, one has that $\phi(x)$ becomes a certain value to implicitly measure the distance between the states $x$ and the target set $\mathcal T$, i.e., $x$ is outside the set $\mathcal T$ if $\phi(x)>0$, and $x$ is in the set $\mathcal T$ if $\phi(x)\leq0$. 
Then, when $x$ is not in the set $\mathcal T$ (i.e., $\phi(x)>0$), we are ready to introduce the minimization of $\phi(x)$ to make $x$ converge to $\mathcal T$,
\begin{align}
\label{normal_execution}
&\min_{u} \phi(x)\nonumber\\
&\mbox{s.t.}~\dot{x}=f(x)+g(x)u
\end{align}
with the control input $u\in\mathbb{R}^q$, and the locally Lipschitz continuous functions of $f(\cdot): \mathbb{R}^d\rightarrow\mathbb{R}^{d},  g(\cdot):\mathbb{R}^d\rightarrow\mathbb{R}^{d\times q}$. The constraint $\dot{x}=f(x)+g(x)u$ in \eqref{normal_execution} denotes a standard dynamic of control affine structure covering large classes of robots~\cite{egerstedt2018robot}. Essentially, $\phi(x)$ decreases along the optimal input $u$ in~\eqref{normal_execution}, which indicates that the robot eventually approaches the target set~$\mathcal T$.

\begin{remark}
The dimension of the state vector $d\in\mathbb{R}^+$ in Eq.~\eqref{normal_execution} may be larger than the dimension of the robot operation $n\in\mathbb{R}^+$ in the Euclidean space (i.e., $d\ge n$), the state vector $x\in\mathbb{R}^d$ in Eq.~\eqref{normal_execution} thus can be utilized to illustrate extra states of robots, such as the orientation, linear velocities and angular velocities of UAVs \cite{mellinger2011minimum}. However, such an illustration is less relevant in this article, because Eq.~\eqref{normal_execution} only provides a generalized optimization paradigm to describe how to make the state vector $x$ converge to the target set $\mathcal T$. 
For the {\it ordering-flexible} target convoying mission in this article, we only need to focus on the $n$-dimensional Euclidean space where the robot operates, rather than the robot's workspace containing additional dimensions,  because the position information $x_i\in\mathbb{R}^n$ with the specific single-integrator dynamics in Eq.~\eqref{kinetic_F} is enough to establish the position-based subtasks in Eqs.~\eqref{ZCBF_1} and \eqref{ZCBF_2} later. For the issue of underactuated characteristics, it has been well addressed by the external low-level tracking module, such as the output regulation control and sliding mode control \cite{liu2019collective}, which is not the main scope of this paper. For the regulation of the robot's orientation, please refer to Remark~\ref{orientation_regulation} for details.

\end{remark}

\begin{remark}
Eq.~\eqref{normal_execution} is solved at each time step and $u^{\ast}$ is the corresponding optimal input. Essentially, $\phi(x)$ in Eq.~\eqref{normal_execution} is regarded as a cost function, and $u^{\ast}$ can be calculated by a typical gradient-descent algorithm~\cite{ruder2016overview} as $u^{\ast}=g^{+}(-f(x)-{\partial \phi(x)}/{\partial x}),$
where $g^{+}\in\mathbb{R}^{q\times d}$ is a Moore-Penrose inverse matrix satisfying $gg^+=I_d$ with the $d$-dimensional identity matrix $I_d\in\mathbb{R}^{d\times d}$. Taking the derivative of $\phi(x)$ along time $t$ and substituting $u^{\ast}$ yields
\begin{align*}
\dot{\phi}(x)%=%&\frac{\partial \phi(x)}{\partial x\t } \dot {x}\nonumber\\
		 =&\frac{\partial \phi(x)}{\partial x\t } \bigg(f(x)+g(x)g^{+}(-f(x)-\frac{\partial \phi(x)}{\partial x})\bigg)\nonumber\\
		 =&-\bigg\|\frac{\partial \phi(x)}{\partial x}\bigg\|^2\leq 0.
\end{align*}
Then, the proper function $\phi(x)$ explicitly decreases along the optimal trajectories when $\|\partial \phi(x)/\partial x\|\neq 0$. In general, convergence to local minima cannot be avoided when $\phi(x)$ is not a convex function.
\end{remark}

However, the traditional task execution in \eqref{normal_execution} only tries the best to make the states $x$ approach the target set $\mathcal T$ because the function $\phi(x)$ is minimized as a cost function, 
which may fail to work any longer in dynamic environments (i.e., $x(T_1)\in\mathcal T\nRightarrow x(t)\in\mathcal T, \forall t>T_1>0$). To rigorously guarantee the task performance for long-term autonomy, a recent work \cite{notomista2021resilient} has proposed an approach of long-term task execution by incorporating $\phi(x)$ into the constraint rather than the cost function.

\begin{definition}
\label{zcbf}
(Long-term task execution)~\cite{notomista2019constraint} For a target set $\mathcal T$ in Eq.~\eqref{task_set}, a robot governed by Eq.~\eqref{normal_execution} 
achieves the long-term task execution if the following constraint-based optimization is solved, i.e.,
\begin{align}
\label{typical_optimization}
&\min_{u, \delta}\|u\|^2+|\delta|^2\nonumber\\
&\mbox{s.t.}~L_f h(x)+L_g h(x)u  \geq\ -\gamma(h(x))-\delta,
\end{align}
where $h(x)=-\phi(x)$ satisfying the inequality in \eqref{typical_optimization} is a control barrier function (CBF) \cite{ames2016control}, $L_f h(x), L_g h(x)$ denote the Lie derivatives of $h(x)$ along the functions $f$ and $g$ in Eq.~\eqref{normal_execution}, respectively, and $\delta\geq 0$ represents a slack variable to measure the violation extent of execution to target set~$\mathcal T$. Here,  
$\gamma(\cdot): (-b,a )\rightarrow (-\infty, +\infty)$ for some $a, b>0$, referred to as the extended class $\mathcal K$ function, is strictly increasing, and satisfies $\gamma(0)=0$, which can determine how fast the states $x$ approaching the target set $\mathcal T$ by setting $\gamma(h(x)):=h(x)^Q$ with different odd positive integer~$Q$~\cite{wang2017safety}. 
\end{definition}

\begin{figure}[!htb]
\centering
\includegraphics[width=6.5cm]{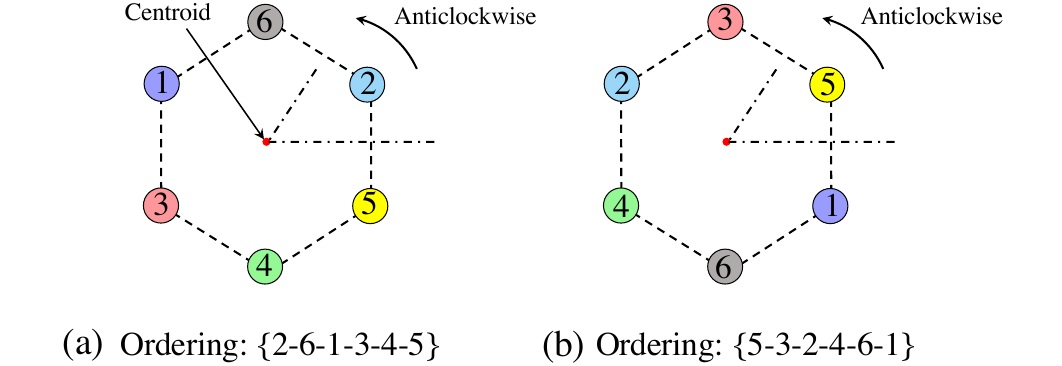}
\caption{Two spatial ordering sequences of a 6-robot hexagonal formation in 2D. (The circles represent the robots, and the red node the centroid.)
}
\label{illustration_ordering}
\end{figure}

In Definition~\ref{zcbf}, by explicitly restricting the state evolution of $h(x):=-\phi(x)$ in the constraint, the long-term task execution \eqref{typical_optimization}  rigorously guarantees advantageous task performance in changing environments. Essentially speaking, the long-term property is ensured because the CBF $h(x)$ endows the forward invariance of the states $x$ to the target set $\mathcal T$ resiliently in a constraint manner \cite{ames2016control}, i.e., 
\begin{align}
\label{forward_invariance}
x(T_1)\in\mathcal T\Rightarrow x(t)\in\mathcal T, \forall t>T_1>0.
\end{align}

\subsection{Ordering-Flexible Target Convoying}
Consider a multi-robot system consisting of $N$ robots denoted by ${\cal V}=\{1,2,\dots, N\} (N\geq3)$, each robot moves according to a single integrator \cite{yao2021singularity}, 
\begin{align}
\label{kinetic_F}
 \dot{x}_i &=u_i, i\in\mathcal V,
\end{align}
where $x_i\in\mathbb{R}^n, u_i\in\mathbb{R}^n$ denote the position and input of robot~$i$ in the $n$-dimensional Euclidean space, respectively. Here, $u_i$ is restricted by a common value~$\zeta\in\mathbb{R}^+$, i.e., $\|u_i\|_{\infty}\leq \zeta$. For the sake of robot generalization, the input $u_i$ in~Eq.~\eqref{kinetic_F} can be regarded as the high-level desired guidance velocity when applied to practical robots of higher-order dynamics. It is applicable to various robots, such as unmanned aerial vehicles (UAVs), autonomous ground vehicles (AGVs) and unmanned surface vessels (USVs)~\cite{zhou2022swarm,hu2023spontaneous,wu2022toward,hu2023coordinated,chen2022synchronization,wei2024resilient}.

\begin{assumption}
\label{assump_tracking}
By regarding $u_i^{\ast}$ of robot $i, i\in\mathcal V$ in Eq.~\eqref{kinetic_F} as the high-level desired velocity, we assume that $u_i^{\ast}$ can be tracked by a low-level velocity-tracking module asymptotically, i.e., $\lim_{t\rightarrow \infty}v_i(t)=u_i^{\ast}$, with $v_i$ being the actual velocity of robot $i$.
\end{assumption}

Assumption~\ref{assump_tracking} is necessary and reasonable in practice because the ROS Controllers\footnote{ROS controller: \href{https://slaterobotics.medium.com/how-to-implement-ros-control-on-a-custom-robot-748b52751f2e}{https://slaterobotics.medium.com/how-to-implement-ros-control-on-a-custom-robot-748b52751f2e}} embedded in many commercially available robotics systems are most likely simple PID controllers, which can only achieve asymptotic convergence.

\begin{figure}[!htb]
\centering
\includegraphics[width=8.0cm]{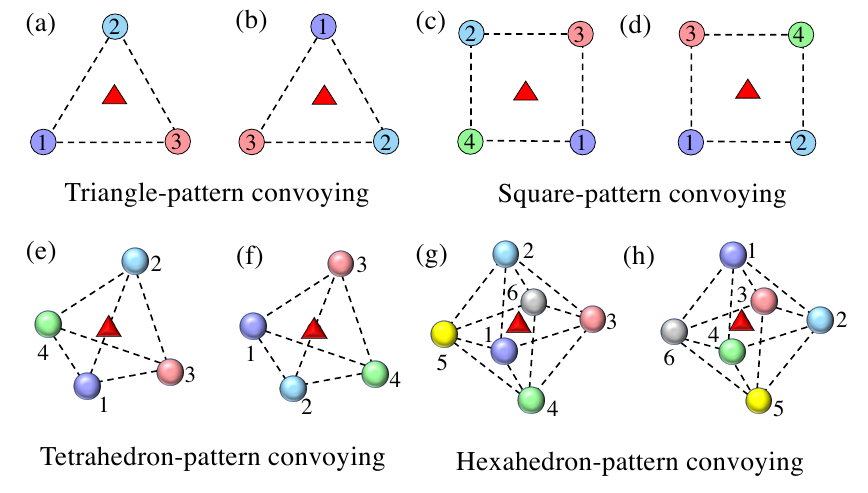}
\caption{ Illustration of four kinds of  {\it ordering-flexible} target convoying in 2D and 3D. (The circles in different colors represent the robots, and the red triangle is the target.)
}
\label{definition_convoying}
\end{figure}

Then, the neighbor set of robot $i$ is defined to be,
\begin{align}
\label{sensing_neighbor}
\mathcal N_i:=\{j\in\mathcal V, j\neq i, \big|~\|x_{i,j}(t)\|\leq R \},
\end{align}
where the collision radius $r\in\mathbb{R}^+$ and the sensing radius $R\in\mathbb{R}^+$ have the same center satisfying $R>r$, and $x_{i,j}:=x_i-x_j$ is the relative position between robots $i$ and $j$. Since $\|x_{i,j}(t)\|$ is time-varying, one has that the robots in the neighbor set $\mathcal N_i$ are time-varying as well, which contribute to {\it ordering-flexible} convoying, but raise challenges in the convergence analysis later.

We also consider a moving target 
\begin{align}
\label{kinetic_target}
\dot{x}_d=v_d,
\end{align}
where $x_d\in\mathbb{R}^n, v_d\in\mathbb{R}^n$ are the position and velocity. From Eq.~\eqref{kinetic_target}, the target can move with a non-constant velocity (i.e., $\dot{v}_d\neq0$), and a constant velocity (i.e., $\dot{v}_d=0$). Note that $\|v_d\|_{\infty}< \zeta$, otherwise the robots cannot chase the moving target.

\begin{assumption}
\label{velocity_estimation}
It is assumed that only the target's position $x_d(t)$ is available to robots but not the velocity $v_d$. Then, there exists a local velocity estimator $\widehat{v}_d^i, i\in\mathcal V,~\|\widehat{v}_d^i\|_{\infty}<\zeta$ for robot $i$ 
satisfying $\lim_{t\rightarrow\infty} \{\widehat{v}_d^i(t)-v_d(t)\}=\mathbf{0}_n$ exponentially.
\end{assumption}

\begin{remark}
\label{target_vel_convergence}
Assumption~\ref{velocity_estimation} illustrates the existence of a local estimator of the moving target velocity $v_d$ using position-only measurements, which is reasonable in practice if each robot is equipped with position sensors.
The design of the local estimator $\widehat{v}_d^i$ in Assumption~\ref{velocity_estimation} has been well studied in literature, e.g., using the output regulation theory \cite{huang2004nonlinear,isidori2013nonlinear,hong2006tracking}. However, there approaches are not main scope of this article, but the vanishing
estimation errors of $\widehat{v}_d^i-v_d$ may influence the convergence of  {\it ordering-flexible} target convoying task. To make the design complete, we give an example of the local estimator for $\dot{v}_d=0$ below,
\begin{align}
\label{example_estimator}
\dot{\widehat{x}}_d^i=&-\chi_1(\widehat{x}_d^i-x_d)+\widehat{v}_d^i,\nonumber\\
\dot{\widehat{v}}_d^i=&-\chi_1\chi_2(\widehat{x}_d^i-x_d),
\end{align}
with the $i$-th estimator $\widehat{\Omega}_i=[\widehat{x}_d^i, \widehat{v}_d^i]\t$ and the estimation gains $\chi_1\in\mathbb{R}^+, \chi_2\in\mathbb{R}^+$. Let $\widetilde{\Omega}_i:=\widehat{\Omega}_i-[x_d, v_d]\t$, one has that $\dot{\widetilde{\Omega}}_i=A\widetilde{\Omega}_i$ with a Hurwitz matrix
\begin{align*}
A=\begin{bmatrix}
-\chi_1  & 1\\
-\chi_1\chi_2 & 0\\
\end{bmatrix},
\end{align*}
which thus implies $\lim_{t\rightarrow\infty}\widetilde{\Omega}_i(t)=\mathbf{0}_{2n}$, i.e., $\lim_{t\rightarrow\infty}\{\widehat{v}_d^i(t)-v_d(t)\}=\mathbf{0}_n$.
\end{remark}

Before presenting the definition of {\it ordering-flexible} target convoying, we introduce the spatial ordering sequence $s[i], i\in\mathbb{Z}_1^N$ of robots in a convex hull first. Notably, there are various rules to determine the spatial ordering. For instance, we can define the 2D spatial ordering starting from the smallest relative angle between the robot and the centroid of all robots, and establish the ordering sequence $s[1], \dots ,s[N]$ anticlockwise around the centroid. An illustrative example is given in Fig.~\ref{illustration_ordering}, where the ordering sequence is determined to be $s[1]=2, s[2]=6, s[3]=1, s[4]=3, s[5]=4, s[6]=5$ in subfigure (a) and $s[1]=5, s[2]=3, s[3]=2, s[4]=4, s[5]=6,$ $ s[6]=1$ in subfigure (b), respectively. Analogously, we can specify the ordering rules for 3D coordination.

\begin{definition}
\label{definition_convoying1}
({\it Ordering-flexible} target convoying)
A multi-robot system $\mathcal V$ governed by Eq.~\eqref{kinetic_F} collectively achieves the {\it ordering-flexible} convoying for a moving target~\eqref{kinetic_target}, if the following three objectives are fulfilled,

1. {\sf\bfseries (Convex-hull convoying)} The robots convoy the moving target into the interior of the convex hull formed by all robots, i.e.,   $\lim\limits_{t\rightarrow\infty}\bigg\{  \sum_{i\in\mathcal V} {x_i(t)}/{N}-x_d(t)\bigg\}=\mathbf{0}_n$ with the position $x_i$ of robot $i$, the number of robots $N$.

2. {\sf\bfseries (Flexible-ordering formation)}  The convoying formation formed by the robots converges to an acceptable region with a steady spatial ordering, i.e.,
\begin{align*}
&(a)~\lim_{t\rightarrow\infty} \big\{\dot{x}_{s[i]}(t)- \dot{x}_{s[i+1]}(t) \big\}=\mathbf{0}_n,\nonumber\\
&(b)~r\leq\lim_{t\rightarrow\infty}\big\|x_{s[i]}(t)-x_{s[i+1]}(t)\big\|<R, \nonumber\\
&\forall i\in\mathbb{Z}_1^{N}~(\mbox{If}~i=N, \mbox{then}~s[i+1]=s[1]),
\end{align*}
where $s[1], s[2], \dots, s[N]$ denote the spatial ordering sequence of robots in a convex hull and are calculated by a bijection $s[i]: \mathbb{Z}_1^N \rightarrow \mathbb{Z}_1^N$ from robot labels $\mathcal V=\{1,\dots, N\}$, $x_{s[i]}$ represents the position of the $s[i]$-th robot, and $R$ is given in \eqref{sensing_neighbor}.

3. {\sf\bfseries (Collision \& Overlapping avoidance)} The inter-robot collision and robot-target overlapping are avoided simultaneously, i.e., 
$\|x_{i,j}(t)\|\geq r, \|x_{i,d}(t)\|>0, \forall i\neq j \in\mathcal V, t>0$
with $x_{i,j}$ given in Eq.~\eqref{sensing_neighbor} and $x_{i,d}:=x_i-x_d$ the relative position between robot $i$ and the target.
\end{definition}

In Definition~\ref{definition_convoying1}, Objective 1) rigorously ensures the basic requirement of convoying the target. The condition~(a) in Objective 2) depicts the rigid formation implicitly, whereas the condition (b) illustrates the key idea of the flexible-ordering formation by re-stipulating steady ordering abide by a specified ordering rule. Note that no spatial ordering is predetermined for specific robots. 
Objective 3) prevents the undesired collision avoidance and singular cases of the overlapping between robots and target), and in turn governs the robots to form the required convex hull eventually. Illustrative examples are shown in Fig.~\ref{definition_convoying}, where the triangle target is convoyed by circle robots with distinct orderings in 2D and 3D Euclidean space.

\subsection{Encoding Long-Term Target Convoying}
Using Definitions~\ref{zcbf} and \ref{definition_convoying1}, we design two subtasks, namely, target approaching and sensing-neighbor collision avoidance to encode the long-term {\it ordering-flexible} target convoying.

It follows from Assumption~\ref{assump_exclusion} that the target approaching subtask $\mathcal T_{i,0}$ of robot~$i$ is described as 
\begin{align}
\label{ZCBF_1}
\phi_{i,0}(x_i)=&\|x_{i}-x_d\|, i\in\mathcal V.
\end{align}
It is observed that $\phi_{i,0}(x_i)$ in \eqref{ZCBF_1} only contains a point $x_d$ in the target set $\mathcal T_0$ and is twice differentiable with the condition of $\|x_{i,d}(t)\|\neq0, \forall t>0$, where such the condition will be guaranteed in Lemma~\ref{lemma_undesired_behavior} later. Then, it follows from the long-term execution in Eq.~\eqref{typical_optimization} and the dynamics in Eqs.~\eqref{kinetic_F}, \eqref{kinetic_target}  that the derivative of $\phi_{i,0}(x_i)$ in Eq.~\eqref{ZCBF_1} satisfies 
\begin{align}
\label{convoying_condition1}
\frac{\partial{\phi}_{i,0}(x_i)}{\partial x_i\t}(u_i-\widehat{v}_d^i)%&=\frac{x_{i,d}\t}{\|x_{i,d}\|}(u_i-\widehat{v}_d^i) \nonumber\\
										&\leq\ -\gamma_{1}( \phi_{i,0}(x_i))+\delta_{i,0},
\end{align}
where $\delta_{i,0}\geq0$ represents the slack variable, $\widehat{v}_d^i$ is the $i$-th local estimator for target's velocity $v_d$ in Assumption~\ref{velocity_estimation}, and $\gamma_1(\psi_{i,0}(x_i))$ is designed to be
\begin{align}
\label{gamma_condition1}
\gamma_1\big(\phi_{i,0}(x_i)\big)=\eta_1\|x_{i,d}\|
\end{align}
with the parameter $\eta_1\in\mathbb{R}^+$ and $x_{i,d}=x_i-x_d$ given in Definition~\ref{definition_convoying1}. Here, $\eta_1\in\mathbb{R}^+$ in Eq.~\eqref{gamma_condition1} is an arbitrary positive constant that will be used to determine the speed at which the robot approaches the target. The larger $\eta_1$ is, the higher the speed $u_i$ of the robot approaching the target will be, until it reaches the maximum input norm $\zeta$, i.e., $\|u_i\|_\infty\leq \zeta$ in (\ref{constrain_convoying_problem}d) later.
The simple formulation of function $\gamma_1(\cdot)$ in Eq.~\eqref{gamma_condition1} will be utilized to prove convex-hull convoying in Lemma~\ref{lemma_convoying} later.

Analogously, the sensing-neighbor collision-free subtask $\mathcal T_{i,j}, j\in\mathcal N_i$ of robot $i$ is designed to be, 
\begin{align}
\label{ZCBF_2}
\phi_{i,j}(x_i)=&r-\|x_{i,j}\|, i\in\mathcal V, j\in\mathcal N_i,
\end{align}
where the sensing neighbors $\mathcal N_i$, the specified collision radius $r\in(0, R)$ are given in \eqref{sensing_neighbor}, and $x_{i,j}$ is given in Eq.~\eqref{sensing_neighbor}. It is observed the target set in Eq.~\eqref{ZCBF_2} is outside the circle $i.e., \|x_{i,j}\|>r$, which is consistent with definition of collision avoidance. Moreover, $\phi_{i,j}(x_i)$ is also twice differentiable for $\|x_{i,j}(t)\|\neq0, \forall t>0$, which will be guaranteed in Lemma~\ref{lemma_undesired_behavior} as well.

%Note that the radius $r$ is stipulated to be larger than actual distance for inter-robot collision avoidance, which will be utilized to eliminate the influence of target-attracting force in Eq.~\eqref{ZCBF_1} later. 

Moreover, from Definition~\ref{zcbf}, one has that $ \phi_{i,j}(x_i), j\in\mathcal N_i,$ in Eq.~\eqref{ZCBF_2} satisfy
\begin{align}
\label{convoying_condition2}
\frac{\partial {\phi}_{i,j}(x_i)}{\partial x_i\t}(u_i-\widehat{v}_d^i)%&{\blue -\frac{x_{i,j}\t} {\|x_{i,j}\|}(u_i-\widehat{v}_d^i) }\nonumber\\
		  \leq& -\gamma_{2}(\phi_{i,j}(x_i)),
\end{align}
where $\gamma_{2}(\phi_{i,j}(x_i))$ is another kind of extended class $\mathcal K$ function 
\begin{align}
\label{gamma_condition2}
\gamma_2\big(\phi_{i,j}(x_i)\big)=\frac{2\zeta \phi_{i,j}(x_i)}{\eta_2}%\frac{e^{\phi_{i,j}}-e^{-\phi_{i,j}}}{e^{\phi_{i,j}}+e^{-\phi_{i,j}}}{\zeta}
\end{align}
with the constant $\eta_2\in\mathbb{R}^+$ and the limit $\zeta$ given in \eqref{kinetic_F}. The function $\gamma_2(\phi_{i,j}(x_i))$ in Eq.~\eqref{gamma_condition2} is designed based on $\zeta$, which will be utilized to prove that the collision-free subtasks for non-neighboring robots are naturally satisfied in Lemma~\ref{lemma_non_neighbor} later. $\eta_2\in \mathbb{R}^+$ in Eq.~\eqref{gamma_condition2} is also an arbitrary positive constant, which will be utilized to determine the minimum of the sensing radius $R$ in Assumption~\ref{assumption_radius} later. More precisely, the larger $\eta_2$ is, the larger the sensing radius $R$ will be. Otherwise, the collision-free constraints for non-neighboring robots in Lemma~\ref{lemma_non_neighbor} may not be satisfied.

Inspired by the long-term task execution in Eq.~\eqref{typical_optimization}, 
it follows from Eqs.~\eqref{ZCBF_1}, \eqref{convoying_condition1}, \eqref{ZCBF_2} and \eqref{convoying_condition2}  that multi-robot long-term target convoying is formulated as a decentralized constraint-based optimization problem for robot $i$ below,
\begin{subequations}
\label{constrain_convoying_problem}
\begin{align}
\min\limits_{u_i, \delta_{i,0}}&\Big\{\|u_i-\widehat{v}_d^i\|^2+l\delta_{i,0}^2\Big\}\\
\mbox{s.t.}~&\frac{\partial \phi_{i,0}}{\partial x_i\t}(u_i-\widehat{v}_d^i)+ \gamma_1(\phi_{i,0})- \delta_{i,0} \leq 0, \\
&\frac{\partial \phi_{i,j}}{\partial x_i\t}(u_i-\widehat{v}_d^i)+ \gamma_2(\phi_{i,j}) \leq 0, \;j\in\mathcal N_i,\\
&\|u_i\|_{\infty}\leq \zeta, \forall i \in\mathcal V,
\end{align}
\end{subequations}
where the position $x_i$, the input $u_i$, the $i$-th velocity estimator $\widehat{v}_d^i$ and the slack varibale $\delta_{i,0}\in\mathbb{R}^+$ are given in Eqs.~\eqref{kinetic_F} and \eqref{convoying_condition1}, and $l\in\mathbb{R}^{+}$ is the corresponding weight. For conciseness, the functions $\phi_{i,0}=\phi_{i,0}(x_i), \phi_{i,j}= \phi_{i,j}(x_i), \gamma_1(\phi_{i,0})=\gamma_1(\phi_{i,0}(x_{i}))$ and $\gamma_2(\phi_{i,j})=\gamma_2(\phi_{i,j}(x_i))$ are given in Eqs.~\eqref{convoying_condition1}, \eqref{convoying_condition2}, \eqref{gamma_condition1} and \eqref{gamma_condition2}, respectively.

The cost function (\ref{constrain_convoying_problem}a) minimizes two items simultaneously, where the minimization of $\|u_i-\widehat{v}_d^i\|^2$ is to track the velocity of the target, and the minimization of $\delta_{i,0}^2$ is to reduce the distance between robot $i$ and the target $x_d$. The weight $l$ is set to be $l>1$, which implies that the reduction of robot-target distance is more important than the velocity tracking. The constraint (\ref{constrain_convoying_problem}b) is the CBF constraint for executing the target-approaching subtask $\mathcal T_{i,0}$ for robot $i$. The introduction of $\delta_{i,0}$ guarantees the feasibility of (\ref{constrain_convoying_problem}b) at the initial time because $x_i(0)$ is not in the target set $\mathcal T_{i,0}$, i.e., $x_i(0)\notin\mathcal T_{i,0}$, otherwise the constraint (\ref{constrain_convoying_problem}b) may not hold at the beginning. The constraint (\ref{constrain_convoying_problem}c) is the CBF constraint for executing the rigorous collision-avoidance subtasks $\mathcal T_{i,j}, j\in\mathcal N_i$ between robots $i$ and $j$. Since there exist no slack variables, one has that $\|x_{i,j}\|\ge r$ will be rigorously guaranteed. The constraint (\ref{constrain_convoying_problem}d) that ensures the control input is bounded by the limit.

\begin{remark}
\label{re_feasibility}
For the feasibility of Eq.~\eqref{constrain_convoying_problem}, there always exists a feasible solution 
\begin{align}
\label{feasible_solution}
\{u_i=\widehat{v}_d^i, \delta_{i,0}>0~\mbox{is sufficiently large}\},
\end{align}
which can satisfy all the constraints (\ref{constrain_convoying_problem}b)-(\ref{constrain_convoying_problem}d). Precisely, from Eq.~\eqref{feasible_solution}, one has that the constraint (\ref{constrain_convoying_problem}b) holds with a sufficiently large $\delta_{i,0}$. 
Due to $\|x_{i,j}(0)\|\ge r$ in Assumption~\ref{assump_distance} later, one has that $\gamma_2({\phi_{i,j}(0)})\le 0$, which implies that the constraints (\ref{constrain_convoying_problem}c) are satisfied by forward-invariance property in Lemma~\ref{lemma_undesired_behavior} later. The constraint (\ref{constrain_convoying_problem}d)
naturally holds due to $\|\widehat{v}_d^i\|_{\infty}<\zeta$ in Assumption~\ref{velocity_estimation}.
\end{remark}

\begin{remark}
Analogous to the LTTE algorithm in Definition~\ref{zcbf}, the proposed framework~\eqref{constrain_convoying_problem} concentrates on minimizing the energy consumption of tracking the velocity of the target in the cost function. Moreover, by encoding time-varying target-convoying subtasks into the rigorous constraints (\ref{constrain_convoying_problem}b)-(\ref{constrain_convoying_problem}c), the long-term property in Eq.~\eqref{constrain_convoying_problem} accounts for the ordering flexibility to tackle changing environmental elements, such as different initial positions, some robots suddenly breakdown, and static obstacles, which will be demonstrated by experiments and simulations in Section~\ref{sec_algorithm_veri} later.
\end{remark}

Now, we are ready to introduce the main problem addressed by this paper.

{\sf\bfseries Problem 1} (Long-term target convoying):
Calculate a coordinated optimized controller
\begin{align}
\label{pro_desired_signal}
[u_{i}\t, \delta_{i,0}]\t:=&\varpi\big(\phi_{i,0},\gamma_1(\phi_{i,0}), \nonumber\\
			   &\phi_{i,j},  \dots, \gamma_{2}(\phi_{i,j})\big), i\in\mathcal V, j\in\mathcal N_i,
\end{align}
for the constraint-based optimization problem~\eqref{constrain_convoying_problem} to achieve multi-robot long-term {\it ordering-flexible} target convoying in changing environments. Here, $\varpi(\cdot): \mathbb{R}\rightarrow\mathbb{R}^{n+1}$ is an unknown implicit function.

\section{Main Results}
In this section, we present the technical results concerning Objectives 1-3 of {\it ordering-flexible} convoying in Definition~\ref{definition_convoying1}, because such three objectives are not intuitive from
the long-term target convoying in \eqref{constrain_convoying_problem} directly.

Before formulating the detailed analysis, we need to guarantee collision-free constraints between non-neighboring robots (i.e., $j\notin \mathcal N_i~\big|~\|x_{i,j}\|>R$) firstly, which, albeit not explicitly  
exhibited in \eqref{constrain_convoying_problem}, cannot be ignored due to the discontinuity of ${\partial \phi_{i,j}}/{\partial x_i\t}$ if $\|x_{i,j}\|>R$.

\begin{assumption}
\label{assumption_radius}
The sensing radius $R$ is set to be larger than a specified value, i.e.,  $R\geq r+\eta_2$ 
with $r$ given in Eq.~\eqref{ZCBF_2} and $\eta_2\in\mathbb{R}^+$ being a positive parameter.
\end{assumption}

Assumption~\ref{assumption_radius} illustrates a minimal sensing radius, which will be utilized to guarantee the collision-free constraints for non-neighboring robots in Lemma~\ref{lemma_non_neighbor}.

\begin{lemma}
\label{lemma_non_neighbor}
Under Assumption~\ref{assumption_radius}, all robots~$\mathcal V$ governed by \eqref{kinetic_F} naturally guarantee the collision-free constraints \eqref{convoying_condition2} for non-neighboring robots with arbitrary input, i.e.,
\begin{align*}
& -\frac{x_{i,j}\t}{\|x_{i,j}\|}(u_i-\widehat{v}_d^i) \leq\ -\gamma_{j}(\phi_{i,j}),\nonumber\\
&  \forall \|u_i\|_{\infty}\leq\zeta,  \|\widehat{v}_d^i\|_{\infty}\leq\zeta, i\in\mathcal V,  j\notin\mathcal N_i,
\end{align*}
with $\eta_2\in\mathbb{R}^+$ given in Eq.~\eqref{gamma_condition2}.
\end{lemma}
 {\it Proof of Lemma~\ref{lemma_non_neighbor}.} 
For arbitrary non-neighboring two robots $i\neq j \in\mathcal V$ satisfying $\|x_{i,j}\|>R$, it follows from the definition of $\phi_{i,j}$ in Eq.\eqref{ZCBF_2} and Assumption~\ref{assumption_radius} that
\begin{align}
\label{h_radius_R}
 \phi_{i,j}=r-\|x_{i,j}\|\leq-\eta_2<0.
%-\frac{1}{2}\ln\frac{1+\eta_3}{1-\eta_3}<0.
\end{align}
Using $\gamma_2(\phi_{i,j})$ in Eq.~\eqref{gamma_condition2}, it follows from Eq.~\eqref{h_radius_R} that 
\begin{align}
\label{h_equality1}
%\frac{e^{\phi_{i,j}}-e^{-\phi_{i,j}}}{e^{\phi_{i,j}}+e^{-\phi_{i,j}}}<-\eta_3<0.
\gamma_2(\phi_{i,j})\leq -2\zeta.
\end{align}
Meanwhile,  the constraints in Eq.~\eqref{convoying_condition2} satisfies
\begin{align}
\label{dh_equality1}
\frac{\partial {\phi}_{i,j}}{\partial x_i\t}(u_i-\widehat{v}_d^i)=-\frac{x_{i,j}\t}{\|x_{i,j}\|}(u_i-\widehat{v}_d^i)\leq2\zeta
\end{align}
with arbitrary inputs $\|u_i\|_{\infty}\leq\zeta,  \|\widehat{v}_d^i\|_{\infty}\leq\zeta$. Combining Eqs.~\eqref{h_equality1} and \eqref{dh_equality1} together yields $ ({\partial \phi_{i,j}}/{\partial x_i\t})(u_i-\widehat{v}_d^i)<-\gamma_2(\phi_{i,j}),$
which implies the collision-free constraints in Eq.~\eqref{convoying_condition2} are naturally satisfied if $\|x_{i,j}\|>R$. The proof is thus completed. 
\eproof

\begin{remark}
The elimination of non-neighboring collis -ion-free constraints in Lemma~\ref{lemma_non_neighbor} not only prevents the complex analysis of discontinuous situation when the robots just enter the sensing radius of another robot $(i.e., \|x_{i,j}\|=R)$, but also features local sensing-only constraints, which can reduce the communication costs and is thus reasonable and economic in practice \cite{wang2017safety}. 
\end{remark}

Based on Lemma~\ref{lemma_non_neighbor}, we can focus on collision-free constraints in \eqref{constrain_convoying_problem} where $\|x_{i,j}\|\leq R$. However, there may still exist singular cases of inter-robot collisions and robot-target overlapping (i.e., $\|x_{i,j}(t)\|< r$ or $\|x_{i,d}(t)\|=0, \exists i\in\mathcal V,  j\in\mathcal N_i$), which makes the constraint-based optimization \eqref{constrain_convoying_problem}  unfeasible. Accordingly, we will firstly prevent the undesired collisions and overlapping (i.e., $\|x_{i,j}(t)\|\ge r, \|x_{i,d}(t)\|>0, \forall i\neq j \in\mathcal V, t>0$), and then prove convex-hull convoying and flexible-ordering formation in Lemmas~\ref{lemma_undesired_behavior}-\ref{lemma_ordering_pattern}, respectively.

\begin{assumption}
\label{assump_distance}
The initial values of inter-robot and robot-target distances are assumed to satisfy
 $\|x_{i,j}(0)\|\geq r$ and $\|x_{i,d}(0)\|>0, \forall i\neq j\in\mathcal V$.
\end{assumption}

Assumption~\ref{assump_distance} is necessary to prevent the undesired situations, otherwise the constraint-based optimization \eqref{constrain_convoying_problem} has no feasible solution at the beginning.

\begin{assumption}
\label{assumption_local_Lipschiz}
The optimal inputs $u_i^{\ast}$ in the constraint-based optimization \eqref{constrain_convoying_problem} are assumed to be locally Lipschitz continuous with respect to its arguments in the target set $\mathcal T_{i,j}$ of Eq.~\eqref{ZCBF_2}.
\end{assumption}
Assumption~\ref{assumption_local_Lipschiz} is a vital condition to prove inter-robot collision avoidance via forward invariance~\cite{ames2016control} in Lemma~\ref{lemma_undesired_behavior} later, because the local Lipschitz continuous property of $u_i^{\ast}$ in \eqref{constrain_convoying_problem} is not assured with the additional input constraints (\ref{constrain_convoying_problem}d), which is different from the traditional QP-based problems (see comment below Eq.~(42) in~\cite{ames2016control}).

\begin{lemma}
\label{lemma_undesired_behavior}
Under Assumptions~\ref{assump_distance} and~\ref{assumption_local_Lipschiz}, all robots $\mathcal V$ governed by the long-term target convoying~\eqref{constrain_convoying_problem} prevent the inter-robot collision avoidance, and robot-target overlapping, i.e., $\|x_{i,j}(t)\|\geq r, \|x_{i,d}(t)\|>0, \forall i\neq j \in\mathcal V, t>0$.
\end{lemma}

{\it Proof.} 
See Appendix~A in Sec~\ref{Proof_lemma_1}.
\eproof

\begin{remark}
When the common global frame is unavailable for robots, it is interesting to see the velocity estimation $\widehat{v}_d^i$ in Remark~\ref{target_vel_convergence} and the long-term target convoying in Eq.~\eqref{constrain_convoying_problem} are still applicable. Precisely, suppose that robot $i$ has its local frame which is the global frame rotated by the matrix $\mathcal R_i\in\mathbb{R}^{n\times n}$ and translated by the position $p_i\in\mathbb{R}^{n}$. The corresponding states in the local frame of robot $i$ are $(\widehat{x}_d^i)^{[i]}=\mathcal R_i\widehat{x}_d^i+p_i, (\widehat{v}_d^i)^{[i]}=\mathcal R_i\widehat{v}_d^i, x_d^{[i]}=\mathcal R_ix_d+p_i, v_d^{[i]}=\mathcal R_i v_d, x_l^{[i]}=\mathcal R_i x_l+p_i, l\in\mathbb{Z}_1^N, u_i^{[i]}=\mathcal R_i u_i$,
where the superscript $[i]$ denotes the $i$-th local frame. Then, the example of velocity estimator in Eq.~\eqref{example_estimator} becomes,
\begin{align*}
(\dot{\widehat{x}}_d^i)^{[i]}=&-\chi_1((\widehat{x}_d^i)^{[1]}-x_d^{[1]})+(\widehat{v}_d^i)^{[i]},\nonumber\\
(\dot{\widehat{v}}_d^i)^{[i]}=&-\chi_1\chi_2((\widehat{x}_d^i)^{[i]}-x_d^{[i]}),
\end{align*}
which also achieves that $\lim_{t\rightarrow\infty}\{(\widehat{v}_d^i)^{[i]}(t)-v_d^{[i]}(t)\}=\mathbf{0}_n$, exponentially. It follow from Eqs.~\eqref{ZCBF_1}, \eqref{gamma_condition1}, \eqref{ZCBF_2} and \eqref{gamma_condition2} that $\|u_i-v_d^i\|=\|u_i^{[i]}-(\widehat{v}_d^i)^{[i]}\|, \phi_{i,0}=\phi_{i,0}^{[i]}, \phi_{i,j}=\phi_{i,j}^{[i]}$,
which implies that the proposed framework in Eq.~\eqref{constrain_convoying_problem} can be implemented in the $i$-th local frame, i.e., 
\begin{subequations}
%\label{local_constrain_convoying_problem}
\begin{align*}
\min\limits_{u_i^{[i]}, \delta_{i,0}}&  \Big\{\|u_i^{[i]}-(\widehat{v}_d^i)^{[i]}\|^2+l\delta_{i,0}^2\Big\}\\
\mbox{s.t.}~&\frac{\partial \phi_{i,0}^{[i]}}{\partial (x_i^{[i]})\t}(u_i^{[i]}-(\widehat{v}_d^i)^{[i]})+ \gamma_1(\phi_{i,0}^{[i]})- \delta_{i,0} \leq 0, \\
&\frac{\partial \phi_{i,j}^{[i]}}{\partial (x_i^{[i]})\t}(u_i^{[i]}-(\widehat{v}_d^i)^{[i]})+ \gamma_2(\phi_{i,j}^{[i]}) \leq 0, \;j\in\mathcal N_i,\\
&\|u_i^{[i]}\|_{\infty}\leq \zeta, \forall i \in\mathcal V.
\end{align*}
\end{subequations}
\end{remark}

\begin{lemma}
\label{lemma_convoying}
Under Assumption~\ref{assump_exclusion}, all robots~$\mathcal V$ governed by the long-term target convoying~\eqref{constrain_convoying_problem} achieve the convex-hull convoying, i.e., 
$\lim\limits_{t\rightarrow\infty}\big\{\sum_{i=1}^Nx_i(t)/N-x_d(t)\big\}=\mathbf{0}_n$.
\end{lemma}

{\it Proof.} 
 See Appendix~B in Sec. \ref{Proof_lemma_2}.
\eproof

\begin{lemma}
\label{lemma_ordering_pattern}
Under Assumption~\ref{assump_distance}, all robots~$\mathcal V$ governed by the long-term target convoying~\eqref{constrain_convoying_problem} form a flexible-ordering formation, i.e., 
\begin{align}
\label{condition_pattern}
&(a)~\lim_{t\rightarrow\infty} \dot{x}_{s[i], s[i+1]}(t) =\mathbf{0}_n,\nonumber\\
&(b)~0<\lim_{t\rightarrow\infty}\|x_{s[i], s[i+1]}(t)\|<R, \nonumber\\
&\forall i\in\mathbb{Z}_1^{N}~(\mbox{If}~i=N, \mbox{then}~s[i+1]=s[1]),
\end{align}
where $x_{s[i], s[i+1]}:=x_{s[i]}-x_{s[i+1]}$.
\end{lemma}

\begin{remark}
\label{balance_remark}
As shown in the optimization problem~\eqref{constrain_convoying_problem}, the time-varying neighbor set $\mathcal N_i(t)$ renders the collision-free constraint (\ref{constrain_convoying_problem}c) time-varying, which implies that the neighboring-robot repulsion provided by the constraints (\ref{constrain_convoying_problem}c) are time-varying as well. Meanwhile, combining with the target-robot attraction from the constraint (\ref{constrain_convoying_problem}b), one has that the constraints (\ref{constrain_convoying_problem}b) and (\ref{constrain_convoying_problem}c) will reach a dynamic balance like physical forces, which finally results in a flexible solution of the optimization problem \eqref{constrain_convoying_problem},  i.e., {\it ordering-flexible} convoying is achieved.
\end{remark}

{\it Proof.} 
See Appendix C in Sec.~\ref{Proof_lemma_3}.
\eproof

\begin{theorem}
\label{theo_convoying}
Under Assumptions~\ref{assump_exclusion}-\ref{assumption_local_Lipschiz}, a multi-robot system in \eqref{kinetic_F} governed by the constraint-based optimization problem~\eqref{constrain_convoying_problem} 
achieves the long-term {\it ordering-flexible} target convoying.
\end{theorem}
{\it Proof of Theorem~\ref{theo_convoying}.}
We draw the conclusion from Lemmas~\ref{lemma_non_neighbor}-\ref{lemma_ordering_pattern} directly.
\eproof

\begin{remark}
\label{orientation_regulation}
Despite the CBF constraints in Eqs.~(\ref{constrain_convoying_problem}b)-(\ref{constrain_convoying_problem}c) being established only based on position, we can still regulate the robot's orientation implicitly. One common approach is to transfer the high-level desired velocity $u_i^{\ast}$ in Eq.~\eqref{constrain_convoying_problem}  to the desired attitude, such as the desired yaw angle $\theta_i^{\ast}=\arctan2(u_{i,2}^{\ast}, u_{i,1}^{\ast})$ in 2D plane, and then leverage an additional low-level tracking module to track $\theta_i^{\ast}$~\cite{hu2021distributed1}. In this way, if all robots have convoyed the target with a rigid pattern, i.e., $u_{i}^{\ast}=v_d, i\in\mathcal V$, the orientation of the robots will converge to be the same. For the robots of simple unicycle dynamics, we alternatively utilize near-identity diffeomorphism (NID) \cite{glotfelter2019hybrid} to regulate the robots' orientation in Section~\ref{2D_experi} later. Moreover, as a preliminary exploration of the ordering-flexible target convoying, this paper only focuses on the homogeneous robots. Despite two kinds of AMRs are employed in the experiments in Section~\ref{2D_experi}, the capabilities of tracking desired velocities are almost the same, which thus does not influence the convoying performance. However, it is still challenging to extend the proposed LTTE algorithm~\eqref{constrain_convoying_problem} to heterogeneous multi-robot systems directly due to the strongly different capabilities of robots, which will be explored in future investigations.
\end{remark}

\begin{figure}[!htb]
\centering
\includegraphics[width=6cm]{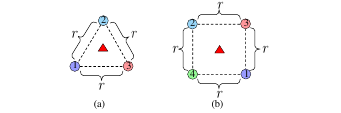}
\caption{ (a) The triangle-pattern convoying is achieved under the undesired equilibria: $\{\|x_{1,2}\|=\|x_{2,3}\|=\|x_{1,3}\|=r$, $ \|x_{i,d}\|$ $>0, i=1,2,3\}.$
(b) The square-pattern convoying is achieved under the undesired equilibria: $\{\|x_{1,4}\|=\|x_{1,3}\|=$ $\|x_{2,4}\|=\|x_{2,3}\|=r, \|x_{i,d}\|>0,$ $i=1,2,3,4\}$.
(All the symbols have the same meaning as in Fig.~\ref{definition_convoying}.)
}
\label{undesired_equilibria}
\end{figure}

\begin{remark}

 The benefits of the proposed framework in Eq.~\eqref{constrain_convoying_problem} is three-fold. (i) Feasibility: By introducing slack variables to the constraints, there always exists a feasible solution for multi-robot long-term target convoying, where details refer to Remark~\ref{re_feasibility}. (ii) Forward invariance: By formulating different target-convoying subtasks into constraints rather than cost functions, Eq.~\eqref{constrain_convoying_problem} inherits forward-invariance property of~Eq~\eqref{typical_optimization} in Definition~\ref{zcbf}, which guarantees rigorous collision avoidance in Lemma~\ref{lemma_undesired_behavior} later. 
(iii) Robustness: By leveraging the time-varying neighbor set $\mathcal N_i(t)$, even if some robots suddenly break down, the rest of robots governed by Eq.~\eqref{constrain_convoying_problem} still find a new solution to achieve the {\it ordering-flexible} target convoying, which is showcased in Fig.~\ref{Broken_3D_circle_trajectory} later.

\end{remark}

\begin{remark}
The  potential undesired equilibria commonly encountered in QP-CLF-CBF and QP-CBF problems, refer to the situation where robots converge to the boundary of the safe set rather than the minimum of 
Lyapunov function~\cite{reis2020control}, (i.e., the undesired equilibrium points in Eq.~\eqref{constrain_convoying_problem} are: $\{\|x_{i,j}\|=r, \|x_{i,d}\|=\delta_{i,0}/\eta_1>0$ and $\delta_{i,0}>0~\mbox{keeps invariant}, i\in\mathcal V, j\in\mathcal N_i\}$ and the constraints (\ref{constrain_convoying_problem}b) and (\ref{constrain_convoying_problem}c) are active). 
However, such undesired equilibria 
still satisfy the condition of the convoying formation, which do not influence the performance of the target-convoying tasks. Specifically,  different from traditional QP-CBF works \cite{notomista2019constraint,notomista2021resilient,reis2020control} approaching the target point (i.e., $\|x_{i,d}\|=0$ in this article), it has been shown in Remark~
\ref{balance_remark} that the proposed optimization problem \eqref{constrain_convoying_problem} is designed to form a convoying formation (i.e., $\sum_{i\in\mathcal V} x_i/N=\mathbf{0}_n, \|x_{i,j}\|\geq r, \|x_{i,d}\|>0$, in 
Definition~\ref{definition_convoying1}) by the balance of the target-approaching and collision-free constraints (\ref{constrain_convoying_problem}b)-(\ref{constrain_convoying_problem}c). 
Therefore, the {\it ordering-flexible} convoying formation can still be formed under the undesired equilibria of $\{\|x_{i,j}\|=r, \|x_{i,d}\|=\delta_{i,0}/\eta_1>0,  \delta_{i,0}>0~\mbox{keeps invariant}, i\in\mathcal V, j\in\mathcal N_i\}$. Illustrative examples are given in Fig.~\ref{undesired_equilibria}, where triangle- and square-pattern convoying are both formed under the undesired equilibria. 
\end{remark}

\begin{figure}[!htb]
\centering
\includegraphics[width=7.5cm]{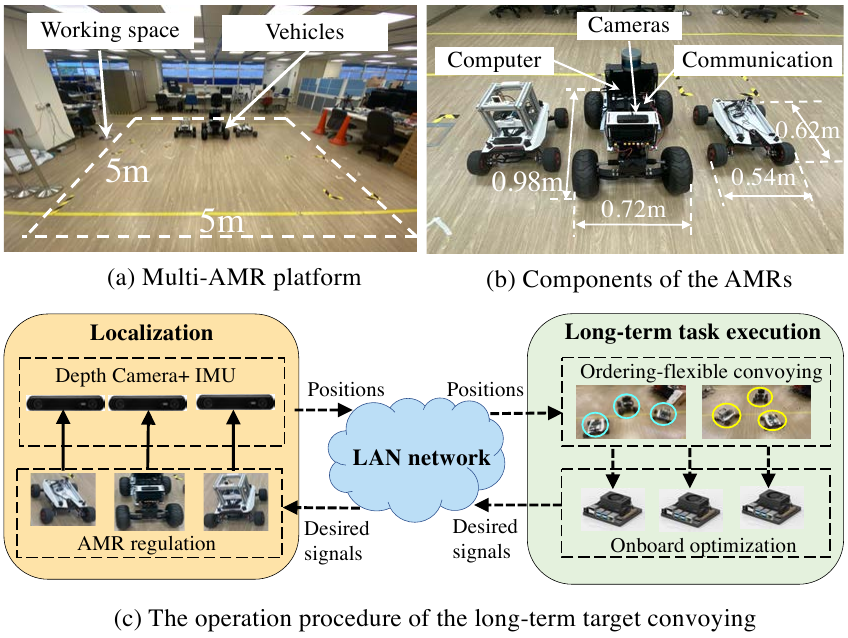}
\caption{ (a) The multi-AMR platform consists of three AMRs and a working space. (b) Sizes of two kinds of AMRs and detailed components. (c) The operation procedure of the target convoying consists of localization, LAN network and long-term task execution. (The solid arrows in subfigure (c) represent the physical connection, and the dashed arrows in subfigure (c) are the virtual connection.)
}
\label{structure}
\end{figure}

\begin{remark}
\label{re_obstacle}
Based on the constraint-driven optimization setup in \eqref{constrain_convoying_problem}, we can further tackle the static obstacle avoidance by seamlessly accommodating the corresponding robot-obstacle subtask $\mathcal T_{i,k}^o,  k\in\mathcal O,$ to be $\phi_{i,k}^o=\|x_{i,k}^o\|-r_k^o$ with $x_{i,k}^o:=x_i-x_k^o$, the position $x_k^o$ and the radius $r_k^o$ to cover the $k$-th circular obstacle in the obstacle set~$\mathcal O$. More precisely, the condition of $\phi_{i,k}^o$ is encoded to be  
\begin{align}
\label{obstacle_condition1}
\frac{\partial {\phi}_{i,k}^o}{\partial x_i\t}(u_i-\widehat{v}_d^i)\leq -\gamma_{2}(\phi_{i,k}^o), i\in\mathcal V,  k\in\mathcal O,  
\end{align}
which can be added into the constraints of the optimization problem~\eqref{constrain_convoying_problem}.  
For the feasibility of Eq.~\eqref{constrain_convoying_problem} with the additional obstacle-avoidance constraints~\eqref{obstacle_condition1}, based on the initial condition of $\phi_{i,k}^o(0)>0$ and forward-invariance property in Eq.~\eqref{forward_invariance}, there also exists a feasible solution $\{u_i=\widehat{v}_d^i, \delta_{i,0}>0~\mbox{is sufficiently large}\},$
%\begin{align}
%\label{feasible_solution_obs}
%\{u_i=\widehat{v}_d^i, \delta_{i,0}>0~\mbox{is sufficiently large}\},
%\end{align}
such that all the constraints (\ref{constrain_convoying_problem}b)-(\ref{constrain_convoying_problem}d), and \eqref{obstacle_condition1} are satisfied, which is similar to Remark~\ref{re_feasibility}. For the undesired local equilibria of Eq.~\eqref{constrain_convoying_problem} with the additional obstacle-avoidance constraints~\eqref{obstacle_condition1}, robots cannot stay at the undesired equilibrium points and will finally leave them. Specifically, after adding the constraints \eqref{obstacle_condition1}, it follows from \cite{reis2020control} that the undesired equilibrium points become: 
%$\{\|x_{i,k}^o\|=r_k^o, \|x_{i,j}\|=r, \|x_{i,d}\|=\delta_{i,0}/\eta_1\neq 0, \delta_{i,0}>0~\mbox{keeps invariant}, i\in\mathcal V, j\in\mathcal N_i\}$ 
\begin{align}
\label{undesired_equilbrium_obs}
\{&\|x_{i,k}^o\|=r_k^o, \|x_{i,j}\|=r, \|x_{i,d}\|=\delta_{i,0}/\eta_1\neq 0,\nonumber\\
 &\mbox{and}~\delta_{i,0}>0~\mbox{keeps invariant}, i\in\mathcal V, j\in\mathcal N_i\}
\end{align}
with the constraints (\ref{constrain_convoying_problem}b), (\ref{constrain_convoying_problem}c) and \eqref{obstacle_condition1} being active. We assume that robots are already at the undesired equilibria when $t=T_3>0$. (i) Since the target $x_d$ is moving with the velocity $v_d$, one has that $x_i$ will moving to satisfy $\|x_{i,d}\|=\delta_{i,0}/\eta_1$ in Eq.~\eqref{undesired_equilbrium_obs}. (ii) However, due to the obstacles are static, it follows from $\|x_{i,k}^o\|=x_i-x_k^o=r_k^o$ in Eq.~\eqref{undesired_equilbrium_obs} that $x_i$ cannot deviate from the circle where the obstacle $x_k^o$ is located, which contradicts (i).
Hence, the undesired equilibrium points will be excluded after some time $t>T_3$. For more complex scenarios of moving obstacles, the feasibility and undesired equilibrium are hard to analyze and will be investigated in future works. 
\end{remark}

\section{Algorithm Verification}
\label{sec_algorithm_veri}
In this section, we will conduct 2D experiments and 3D simulation for algorithm verification.

\begin{figure}[!htb]
\centering
\includegraphics[width=8cm]{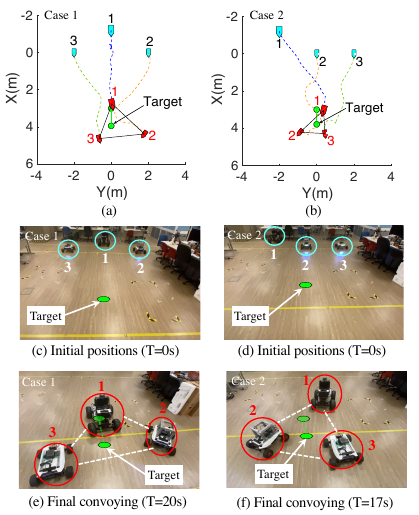}
\caption{ Two experimental cases of {\it ordering-flexible} target convoying with different initial positions. Subfigures (a)-(b): Trajectories of three AMRs starting from different initial position form a convoying formation with distinct spatial orderings in Cases 1-2. Subfigures (c)-(d): Snapshots of initial positions of the three AMRs in Cases 1-2. Subfigures (e)-(f): Snapshots of final stable convoying formation formed by the three AMRs in Cases 1-2.  (Here, the blue and red vehicles in subfigures (a)-(b) denote initial and the final positions of the three AMRs, respectively. The green circle denotes the moving target. Moreover, the lager AMR Hunter $1.0$ is labeled $1$, whereas the other two Scout Mini are specified with labels $2, 3$.)
}
\label{expe_trajectory}
\end{figure}

\begin{figure}[!htb]
\centering
\includegraphics[width=6.9cm]{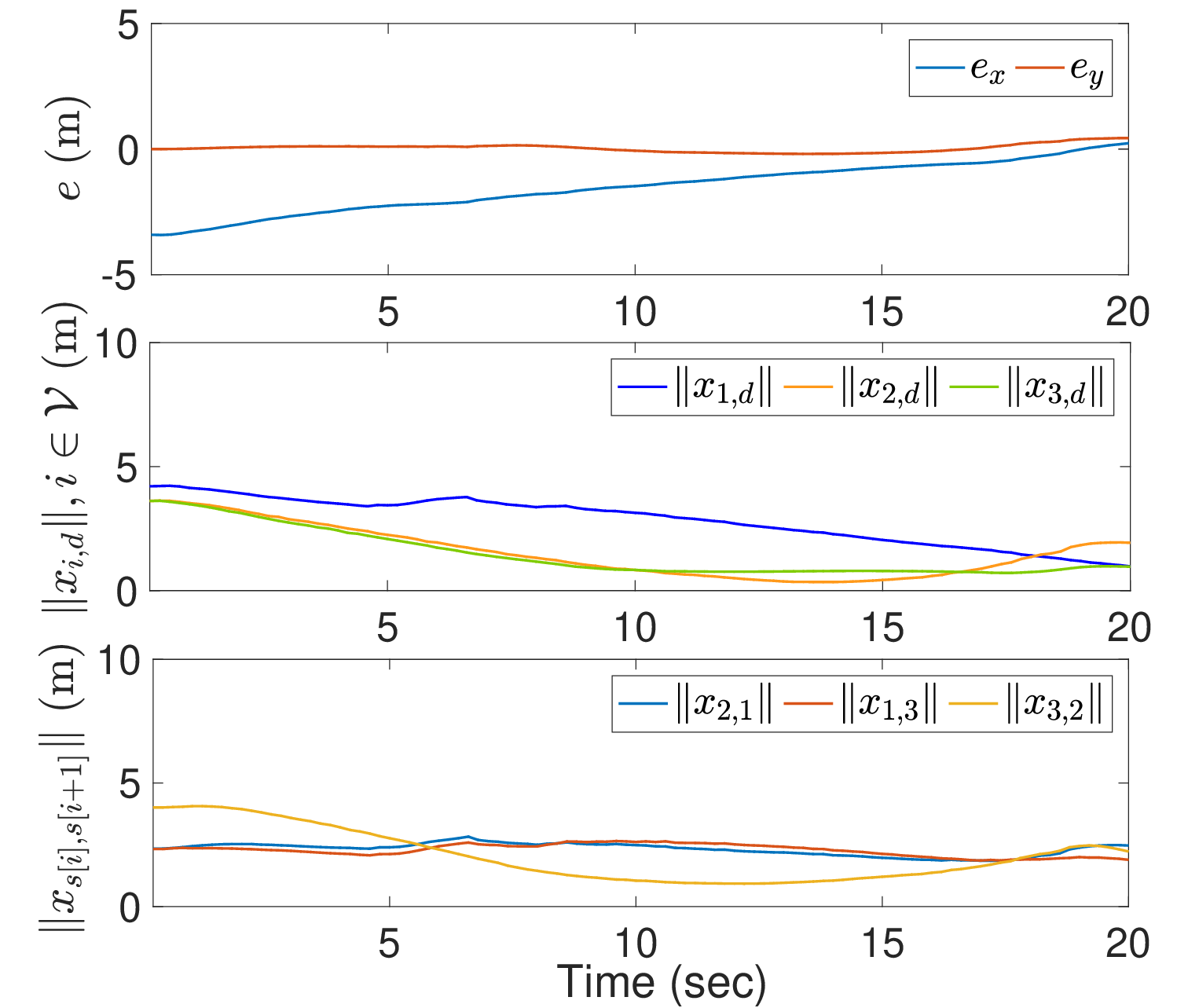}
\caption{ Temporal evolution of the convoying errors $e=[e_{x}, e_y]\t\in\mathbb{R}^2$, the robot-target distances $\|x_{i,d}\|, i=1,2,3,$ and the relative distances between adjacent robots $\|x_{2,1}\|, \|x_{1,3}\|, \|x_{3,2}\|$ with an ordering sequence in a 2D plane in Fig.~\ref{expe_trajectory} (a) for example.
}
\label{2D_exper_distance}
\end{figure}

\begin{figure}[!htb]
\centering
\includegraphics[width=6.9cm]{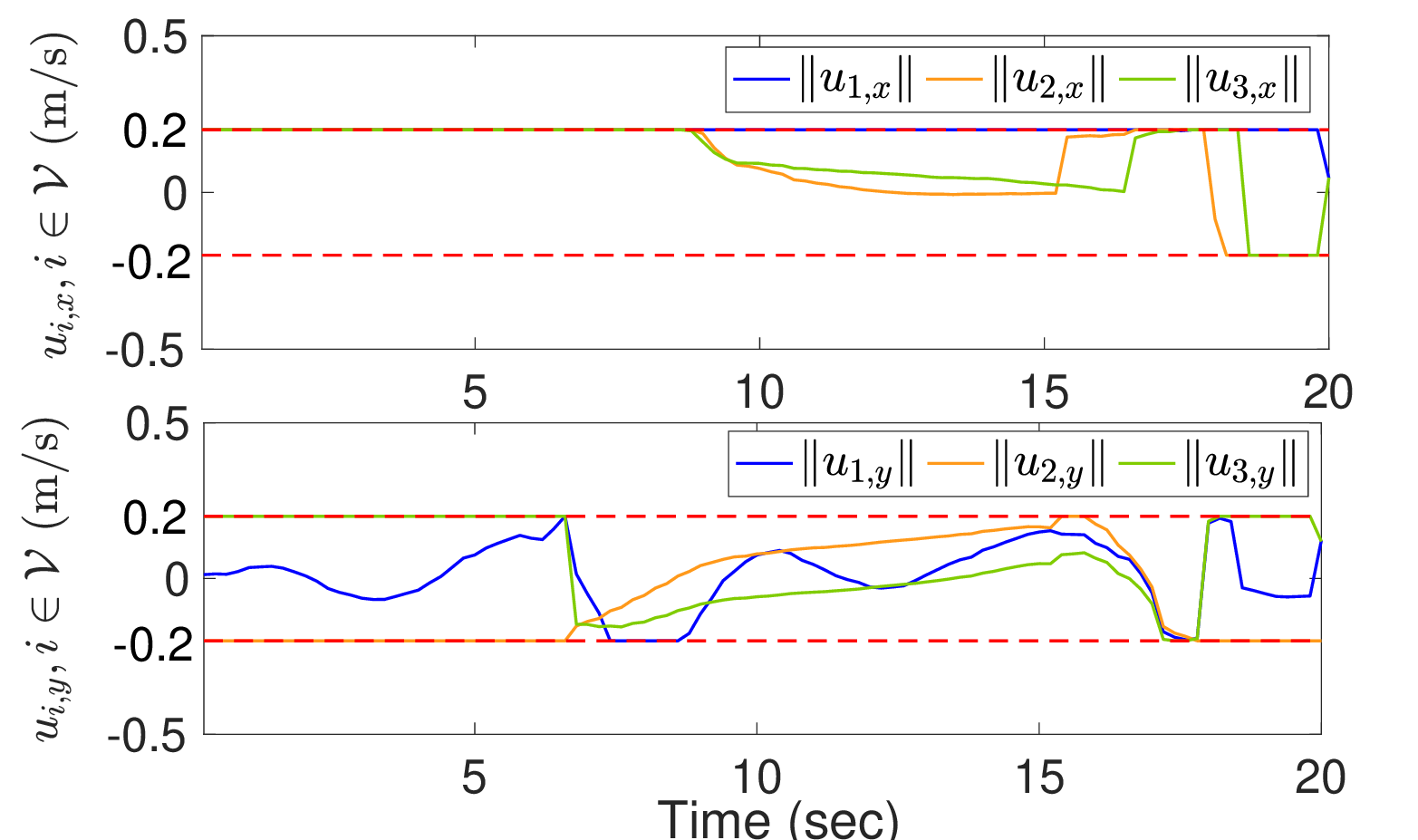}
\caption{ Temporal evolution of the AMR desired velocities $u_i:=[u_{i,x},$ $ u_{i,y}]\t\in\mathbb{R}^2, i=1,2,3,$  in Fig.~\ref{expe_trajectory} (a) for example. (The dashed red lines are the upper and lower limits of the input)
}
\label{2D_exper_input}
\end{figure}

\subsection{2D Experiments}
\label{2D_experi}
In this subsection, we conduct 2D experiments with a multi-AMR platform to verify the effectiveness of long-term {\it ordering-flexible} target convoying \eqref{constrain_convoying_problem}. To proceed, we firstly introduce the multi-AMR platform. As shown in Fig.~\ref{structure} (a), the multi-AMR platform is composed of a $5$m $\times$ $5$m working space and three AMRs, where the biggest one is Hunter $1.0$ and the other two are Scout Mini\footnote{AMRs: \href{https://global.agilex.ai/}{https://global.agilex.ai/}}. It is observed in Fig.~\ref{structure} (b) that the Hunter-$1.0$ AMR is $0.98$m in length and $0.72$m in width, and the Scout Mini AMR is $0.62$m in length and $0.54$m in width, where each AMR is equipped with an onboard computer: NVIDIA Jetson Xavier NX\footnote{NVIDIA Jetson Xavier NX: \href{https://www.nvidia.com/en-us/autonomous-machines/embedded-systems/jetson-xavier-nx/}{https://www.nvidia.com/en-us/autonomous-machines/embedded-systems}}, a depth stereo camera: ZED$2i$\footnote{ZED$2i$: \href{https://www.stereolabs.com/zed-2i/}{https://www.stereolabs.com/zed-2i/}} integrating depth detection and inertial measurement unit (IMU), and a ROS TCP-protocol communication component. Fig.~\ref{structure} (c) illustrates the detailed operation procedure, where the AMR leverages the ZED2i camera to localize its position with a typical VINS-Fusion SLAM algorithm \cite{qin2019general}, and then broadcasts the position information to the onboard computer through a common WiFi LAN network. 
Then, based on the positions from its own and sensing neighbors, the desired signals are calculated by the onboard computer and then sent to the actuators of the AMRs. Notably, the tracking of the desired signals in AMR regulation module has been well achieved by AGILE$\cdot$X company in advance.

To accommodate the optimal $u_i^{\ast}$ in Eq.~\eqref{constrain_convoying_problem} to the commercial AMRs with desired speeds and rotation rates, 
we consider the following unicycle model for AMRs \cite{yao2021singularity}
\begin{align}
\label{unicycle}
\dot{x}_{i,1}=&v_i\cos\theta_i,~\dot{x}_{i,2}=v_i\sin\theta_i,~\dot{\theta}_i=u_{\theta_i}, i\in\mathcal V,
\end{align}
where $x_i=[x_{i,1}, x_{i,2}]\t\in\mathbb{R}^{2}$ is the position, $\theta_i\in \mathbb{R}$ is the yaw angle in X-Y plane, $v_i\in\mathbb{R}, u_{\theta_i}\in\mathbb{R}$ are the corresponding desired velocity and rotation rate of robot~$i$, respectively. According to near-identity diffeomorphism (NID) \cite{glotfelter2019hybrid}, the desired velocities $\{u_{i,1}^{\ast}, u_{i,2}^{\ast}\}$ in Eq.~\eqref{constrain_convoying_problem} can be transformed to the desired commands $\{v_i, u_{\theta_i}\}$ in Eq.~\eqref{unicycle} below
\begin{align*}
v_{i}=&u_{i,1}^{\ast}\cos\theta_i +u_{i,2}^{\ast}\sin\theta_i, \nonumber\\
u_{\theta_i}=&-\frac{u_{i,1}^{\ast}}{l}\sin\theta_i +\frac{u_{i,2}^{\ast}}{l} \cos\theta_i, i\in\mathcal V,
\end{align*}
with $l\in\mathbb{R}^+$ being an arbitrary small constant. In the experiments, $l$ is set to be $l=0.2$ for convenience.

\begin{figure*}[!htb]
\centering
\includegraphics[width=\hsize]{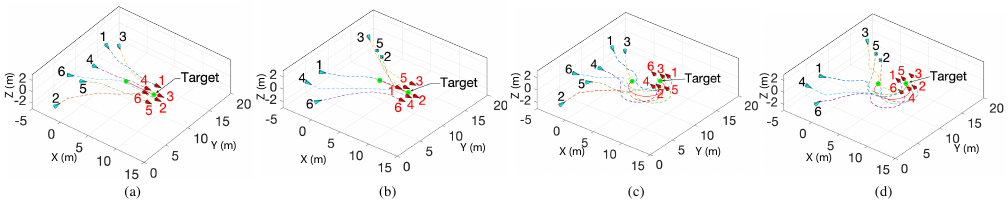}
\caption{ Four cases of a six-robot system from two different initial positions
to the {\it ordering-flexible} hexahedron-pattern convoying for a target of different dynamics using the proposed constraint-based optimization framework~\eqref{constrain_convoying_problem} in 3D. 
Subfigures (a)-(b): A line-motion {\it ordering-flexible} target convoying ( i.e., a constant velocity $v_d:=[1,0,0]\t$ m/s). Subfigures (c)-(d): A circular-motion {\it ordering-flexible} target convoying ( i.e., a variational velocity  $v_d:=[2\omega\cos(\theta+\pi/2), 2\omega\sin(\theta+\pi/2, 0]\t$ m/s) with $\theta$ being the relative angle between the target $x_d$ and the centroid of desired circle, $\omega$ the corresponding angular velocity. (Here, the blue and red triangles represent the initial, and final positions of the robots, respectively. The green ball and red line
denote the target and its trajectory, respectively.)
}
\label{3D_line_trajectory}
\end{figure*}

\begin{figure}[!htb]
\centering
\includegraphics[width=7.0cm]{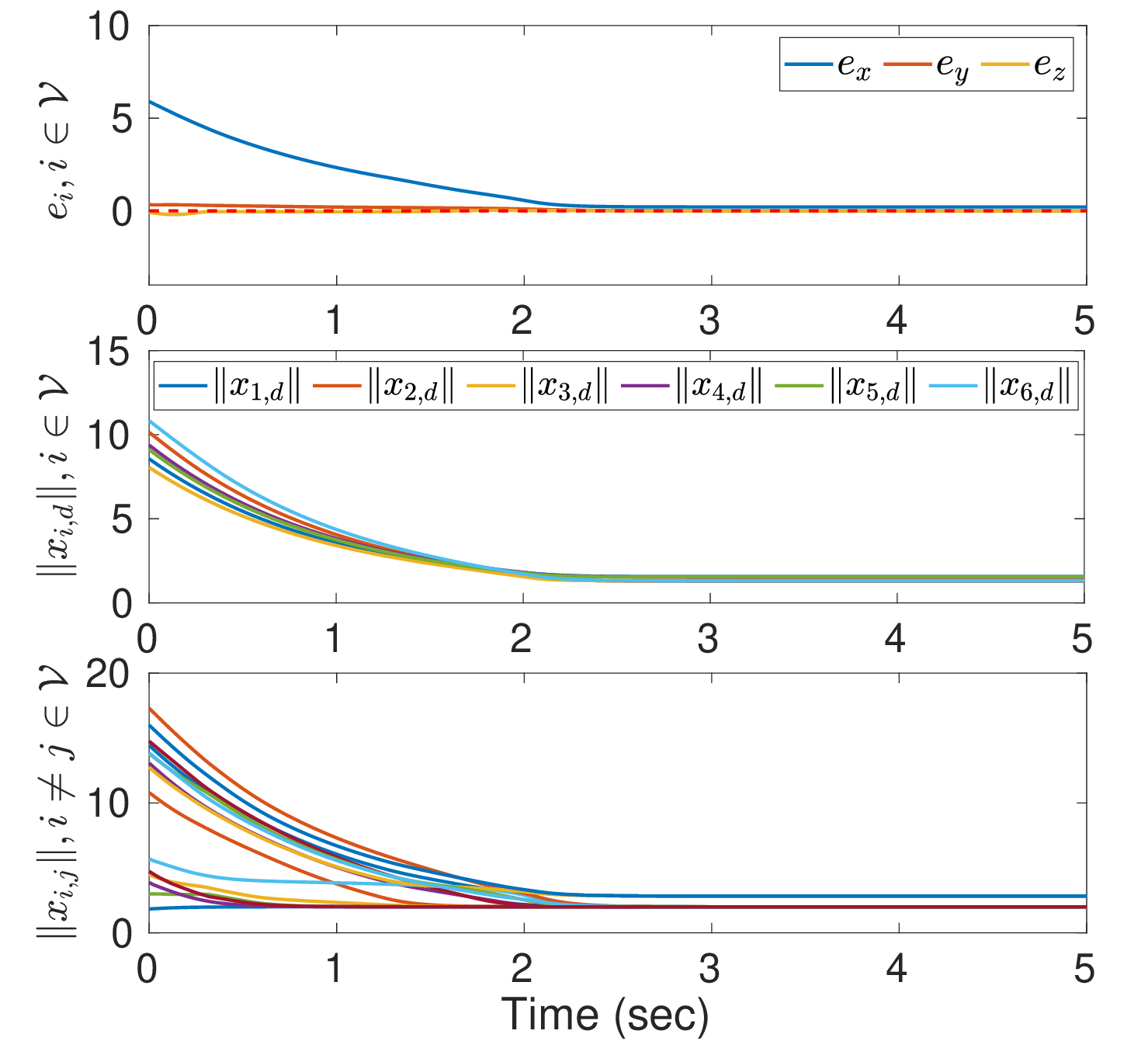}
\caption{ Temporal evolution of the convoying errors $e=[e_{x}, e_y, e_z]\t\in\mathbb{R}^3$, the robot-target distances $\|x_{i,d}\|$, $i\in\mathcal V,$ and the inter-robot distances $\|x_{i,j}\|, i\neq j\in\mathcal V$ in Fig.~\ref{3D_line_trajectory} (b) for example.
}
\label{3D_distance}
\end{figure}

\begin{figure}[!htb]
\centering
\includegraphics[width=7.0cm]{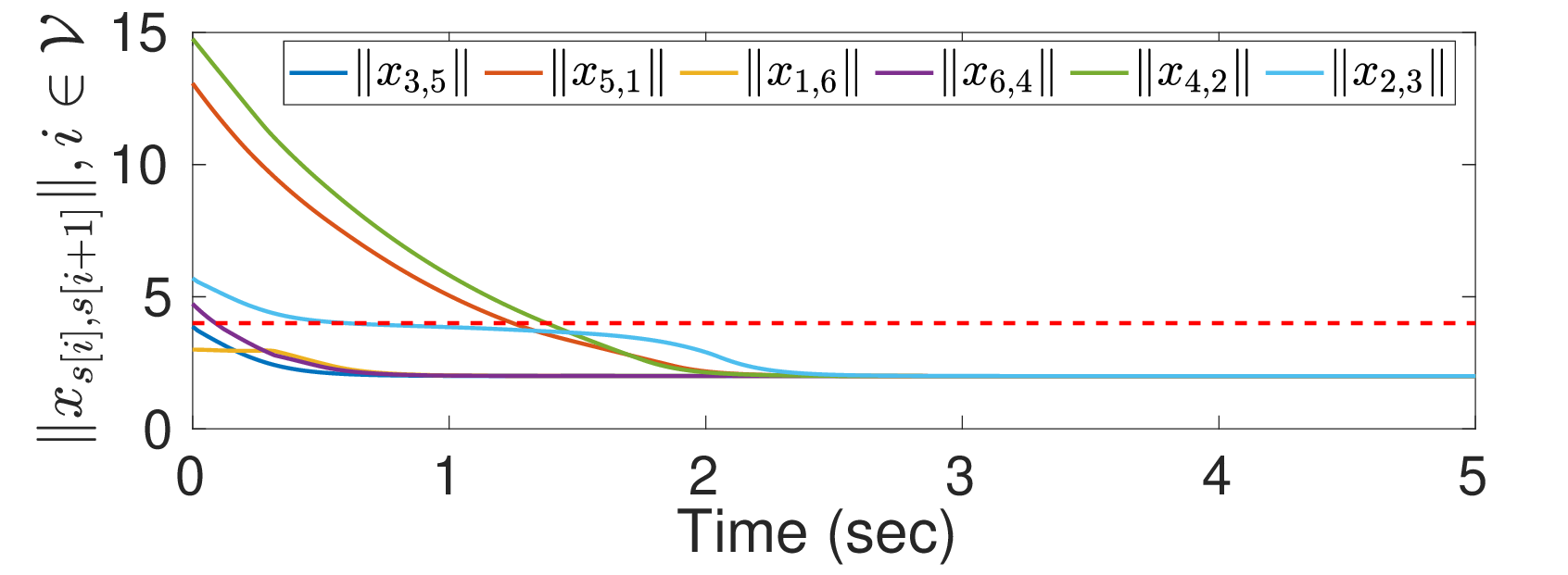}
\caption{ Temporal evolution of the relative distances between adjacent robots in the ordering sequence $\|x_{3,5}\|, \|x_{5,1}\|$, $ \|x_{1,6}\|, \|x_{6,4}\|, \|x_{4,2}\|$ in Fig.~\ref{3D_line_trajectory}~(b) for example.
}
\label{3D_ordering_distance}
\end{figure}

\begin{figure}[!htb]
\centering
\includegraphics[width=7.0cm]{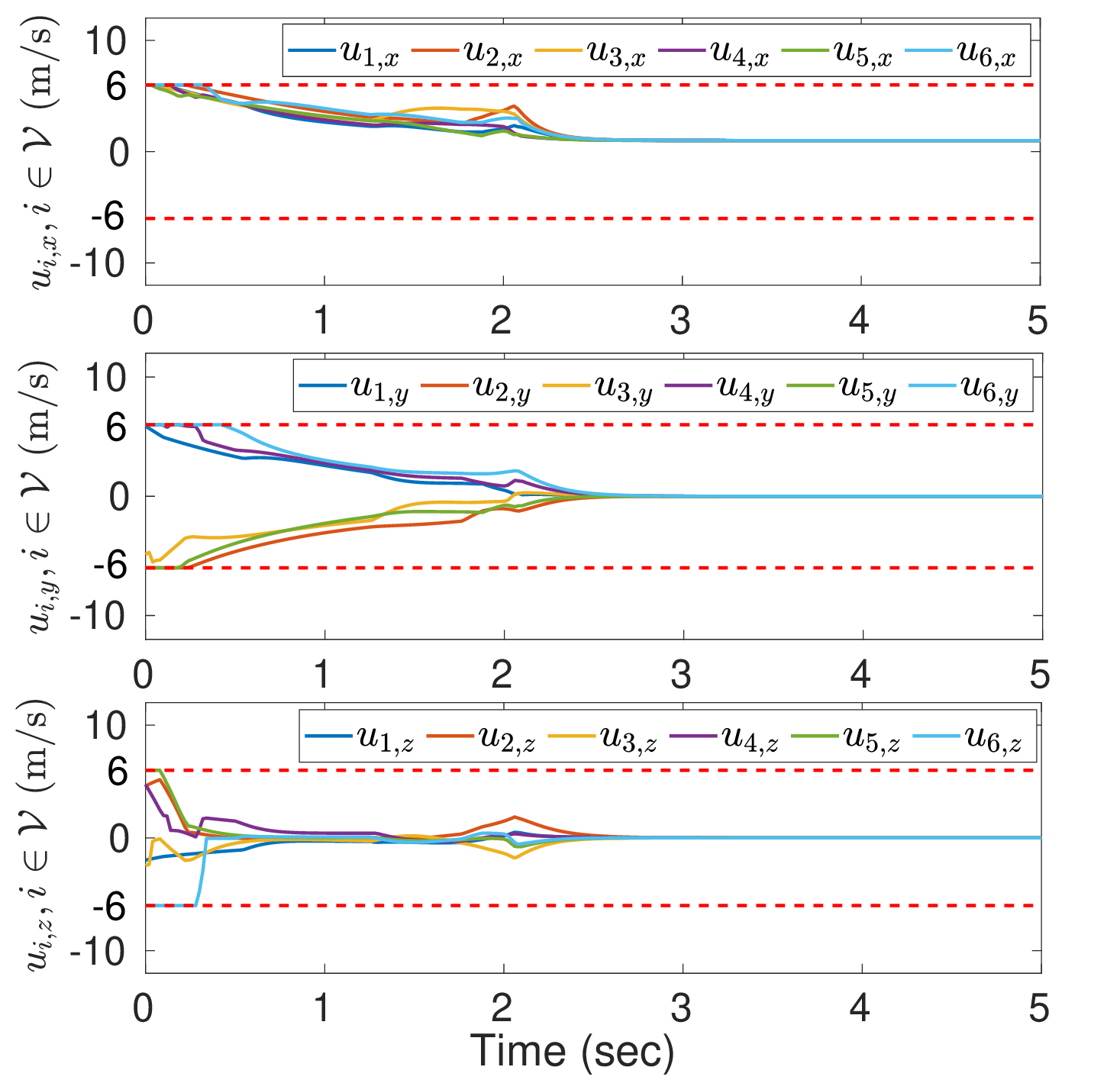}
\caption{ Temporal evolution of the robot inputs $u_i:=[u_{i,x},$ $ u_{i,y},u_{i,z}]\t\in\mathbb{R}^3, i\in\mathcal V,$  in Fig.~\ref{3D_line_trajectory} (b) for example. (The dashed red lines are the upper and lower limits of the input.)
}
\label{3D_inputs}
\end{figure}

In what follows, we consider the {\it ordering-flexible} convoying task with three AMRs and a constant-velocity virtual target because of the limited working space. It follows from Eq.~\eqref{kinetic_F} that the input limit is set to be $\zeta=0.2$, i.e., $\|u_i\|_{\infty}\leq 0.2, i\in\mathbb{Z}_1^3$.
Due to the high velocity bound $\xi$ possibly leading to Scout Mini and Hunter AMRs exceeding the limited working space and behaving with different performances, we set $\xi=0.2$ to be much smaller than the maximum speed of Scout Mini ($2.7$ m/s) and Hunter 1.0 ($4.8$ m/s), which is to make the AMRs maintain turning rate as low as possible at a low speed. Moreover, we set the limit $\xi$ to be the same to endow each AMR with almost the same capabilities. 
Moreover, the constructive procedure of selecting $r, R, \eta_2$ is given below. (i) According to the size of Scout Mini and Hunter AMRs, we determine the collision radius to be $r=1.5$m. (ii) Based on the effective range of the ZED2i camera, we set the sensing radius to be $R=4$m. (iii) It follows from Assumption~\ref{assumption_radius} that $\eta_2=R-r=2.5$. The gain $\eta_1$ in Eq.~\eqref{convoying_condition1} is set to be $\eta_1=2.0$, and the initial value of the slack variable $\delta_{i,0}$ is set to be $\delta_{i,0}(0)=100$. Due to the large size of AMRs and the wide turning radius ($1.6$ m) of Hunter 1.0 AMR occupying much space in limited working space,
the velocity of the virtual target is thus set to be a constant $v_d=[0.06,0]\t$m/s. Then, a distributed velocity estimator with an extra connected communication topology is designed to track the target's velocity according to \cite{hong2006tracking}. Fig.~\ref{expe_trajectory} illustrates two experimental cases  (i.e., Cases 1-2) of the three AMRs collectively forming convoying formation for the moving target. More precisely, as shown in Figs.~\ref{expe_trajectory} (a), (b), three AMRs starting from different initial positions (blue vehicles) fulfilling Assumption~\ref{assump_distance} successfully form a triangle-pattern target convoying (red vehicles) with two distinct orderings $\{2, 1, 3\}$ and $\{3, 1,2\}$ in Cases 1-2, respectively. The corresponding snapshots of initial positions and final-convoying formation of Cases 1-2 are exhibited in Figs.~\ref{expe_trajectory} (c), (d), (e), (f) to illustrate the effectiveness of the long-term target convoying~\eqref{constrain_convoying_problem}.

Additionally, we take Fig.~\ref{expe_trajectory} (a) as an illustrative example to quantitatively analyze the corresponding state evolution. As shown in Fig.~\ref{2D_exper_distance}, the convoying errors $e=[e_x, e_y]\t\in\mathbb{R}^3$ in Eq.~\eqref{converge_convoy_err} converge to be around zeros at $t=20$s, which indicate that the virtual target is eventually convoyed by the three AMRs, i.e., Objective 1) in Definition~\ref{definition_convoying}. The robot-target distances $\|x_{i,d}\|, i=1,2,3,$ and the inter-robot distances in Fig.~\ref{2D_exper_distance} satisfy $\|x_{2,1}(t)\|>1.5, \|x_{1,3}(t)\|>1.5, \|x_{3,2}(t)\|>1.5, \forall t>0$, which avoid the overlapping behavior in Objective 3) of Definition~\ref{definition_convoying}. It is observed in Fig.~\ref{2D_exper_distance} that the relative distance between the adjacent robots along the ordering sequence satisfies $1.5<\lim_{t\rightarrow\infty}\|x_{s[i],s[i+1]}(t)\|<4, s[1]-s[3]:=\{2,1,3 \}$, which ensures Objective 2) of Definition~\ref{definition_convoying}. Moreover, Fig.~\ref{2D_exper_input} depicts that the desired velocities $u_i=[u_{i,x}, u_{i,y} ]\t\in\mathbb{R}^3, i=1,2,3,$ of robot $i$ are bounded by the limit $\zeta=0.2$ all the time, i.e., $\|u_i(t)\|_{\infty}\leq 0.2, \forall t>0$, which guarantee the feasibility of the long-term target convoying \eqref{constrain_convoying_problem}.

\begin{remark}
The velocity-tracking module of the AMRs employed in 2D experiments has been encapsulated by AGILE$\cdot$X Company, which implies that the public users cannot access the detailed tracking methods but only transmit the desired command velocities to AMRs. Since the velocity tracking module provided by AGILE·X Company is not available to public users, we have conducted several velocity tracking tests to ensure its performance before implementing convoying experiments. Precisely, the AMRs can track constant and varying velocities asymptotically, and there only exist some tiny disturbances. Moreover, the influence of such tiny disturbances is further mitigated by the outer-loop target convoying~\eqref{constrain_convoying_problem} via feedback, because the velocity-tracking module is in the inner loop of the closed-loop system. Therefore, it suffices to conduct the ordering-flexible target convoying experiments with the encapsulated velocity-tracking module.
It is still worth mentioning that the proposed framework~\eqref{constrain_convoying_problem} cannot accommodate frequent transient behavior directly when tracking commanded velocities $u_i, i\in\mathcal V$. 
If $u_i, i\in\mathcal V,$ in Eq.~\eqref{constrain_convoying_problem} is transiently changing, one has that the derivative of $u_i$ will become infinite at some time $t=T_f>0$ (i.e., $\lim_{t\rightarrow T_f}\dot{u}_i=\infty$), which contradicts the bounded states $\ddot{\Phi}_i$ and the convergence of convex-hull convoying in Lemma~\ref{lemma_convoying} cannot be guaranteed. Therefore, the special transient behavior will be investigated in future works.
\end{remark}

\begin{figure}[!htb]
\centering
\includegraphics[width=9.5cm]{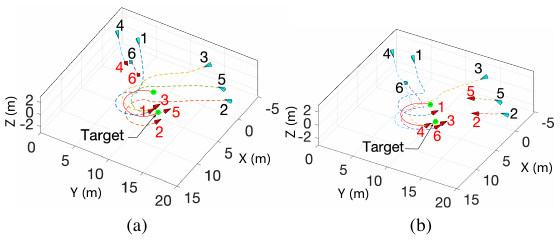}
\caption{ Special situations of two robots suddenly breakdown at $t=0.5$s in the six-robot {\it ordering-flexible} target convoying task. Subfigure (a): The rest four robots $i=1, 2, 3, 5$ form the successful tetrahedron-pattern convoying when robots $i=4, 6$ break down.
Subfigure (b): The rest four robots $i=1, 3, 4, 6$ form the successful tetrahedron-pattern convoying when robots $i=2, 5$ break down. (All the symbols have the same meaning as in Fig.~\ref{3D_line_trajectory}.)
}
\label{Broken_3D_circle_trajectory}
\end{figure}

\subsection{3D Simulations}
\label{3D_simulations}
In this subsection, 3D simulations of {\it ordering-flexible} target convoying are conducted to verify the feasibility of Theorem~\ref{theo_convoying} with changing environmental elements in higher-dimensional Euclidean space.
We consider a multi-robot system of six robots governed by~\eqref{kinetic_F} and \eqref{constrain_convoying_problem} with the input limit specified to be $\zeta_i=6$.  We choose the same collision radius $r$, sensing radius $R$, the parameters $\eta_1, \eta_2$ as in Section \ref{2D_experi} for convenience.
In what follows, we consider the {\it ordering-flexible} multi-robot convoying with a moving target of line motion and circular motion, respectively.

For the line-motion target, we assume that the target moves with a constant velocity $v_d:=[1,0,0]\t$ m/s, which is tracked by the similar velocity estimator in Section \ref{2D_experi}.  Figs.~\ref{3D_line_trajectory} (a)-(b) illustrate the trajectories of six robots from different initial positions (blue
triangles) fulfilling the Assumption~\ref{assump_distance} to the hexahedron-pattern target convoying (red triangles) with two distinct orderings $\{1, 4, 6, 5, 2, 3\}$ and $\{3, 5, 1, 6, 4, 2\}$
 in the 3D Euclidean space. For the circular-motion target, we specify the target's velocity to be $v_d:=[2\omega\cos(\theta+\pi/2), 2\omega\sin(\theta+\pi/2, 0]\t$ m/s, where $\theta$ is the relative angle between the target $x_d$ and the centroid of desired circle $[6,10,0]\t$, and $\omega=0.1s^{-1}$ is the corresponding angular velocity. Similarly, Figs.~\ref{3D_line_trajectory} (c)-(d) describe the
trajectories of six robots from same initial positions setup in Figs.~\ref{3D_line_trajectory} (a)-(b) (blue
triangles) fulfilling the Assumption~\ref{assump_distance} to the desired hexahedron-pattern convoying (red triangles) with distinct orderings $\{1, 3, 6, 4, 2, 5\}$ and $\{3, 5, 1, 6, 4, 2\}$. Both of the previous simulations demonstrate the flexible-ordering property in Objective 2 of Definition~\ref{definition_convoying1}.

\begin{figure}[!htb]
\centering
\includegraphics[width=6.8cm]{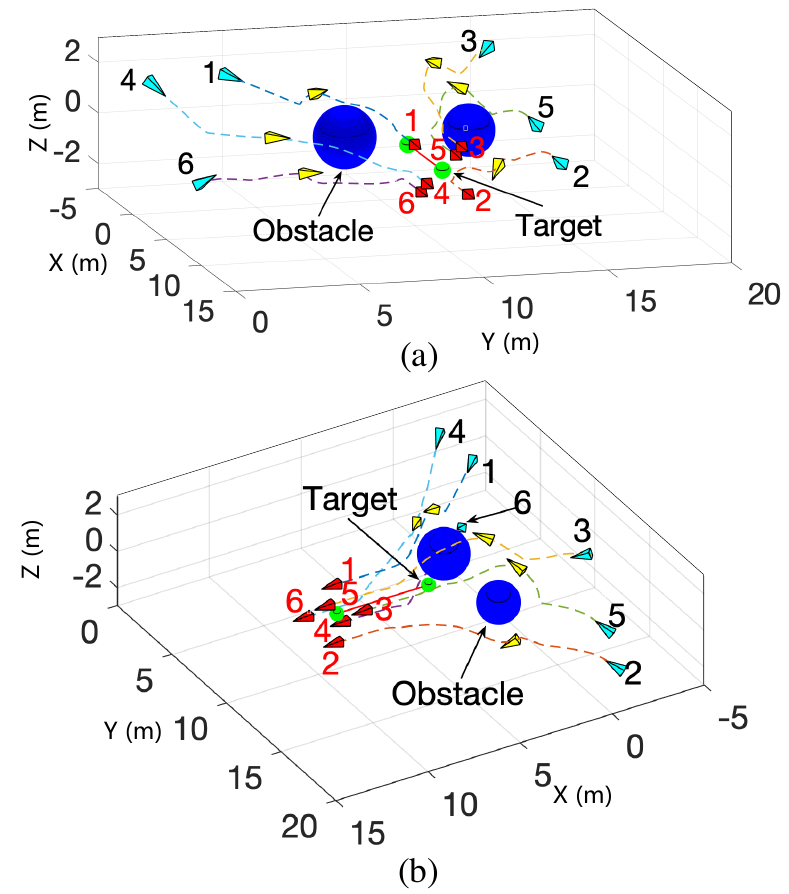}
\caption{ (a) Trajectories of six robots and the target from the same initial positions in Fig.~\ref{3D_line_trajectory} (b) with two static ball obstacles. (b) Another view to better illustrate the desired hexahedron-pattern convoying.
(Here, the yellow triangles
represent the obstacle-encountering positions of robots. All the other symbols have the same meaning as in Fig.~\ref{3D_line_trajectory}.)
}
\label{Obs_3D_line_trajectory}
\end{figure}

\begin{figure}[!htb]
\centering
\includegraphics[width=7.0cm]{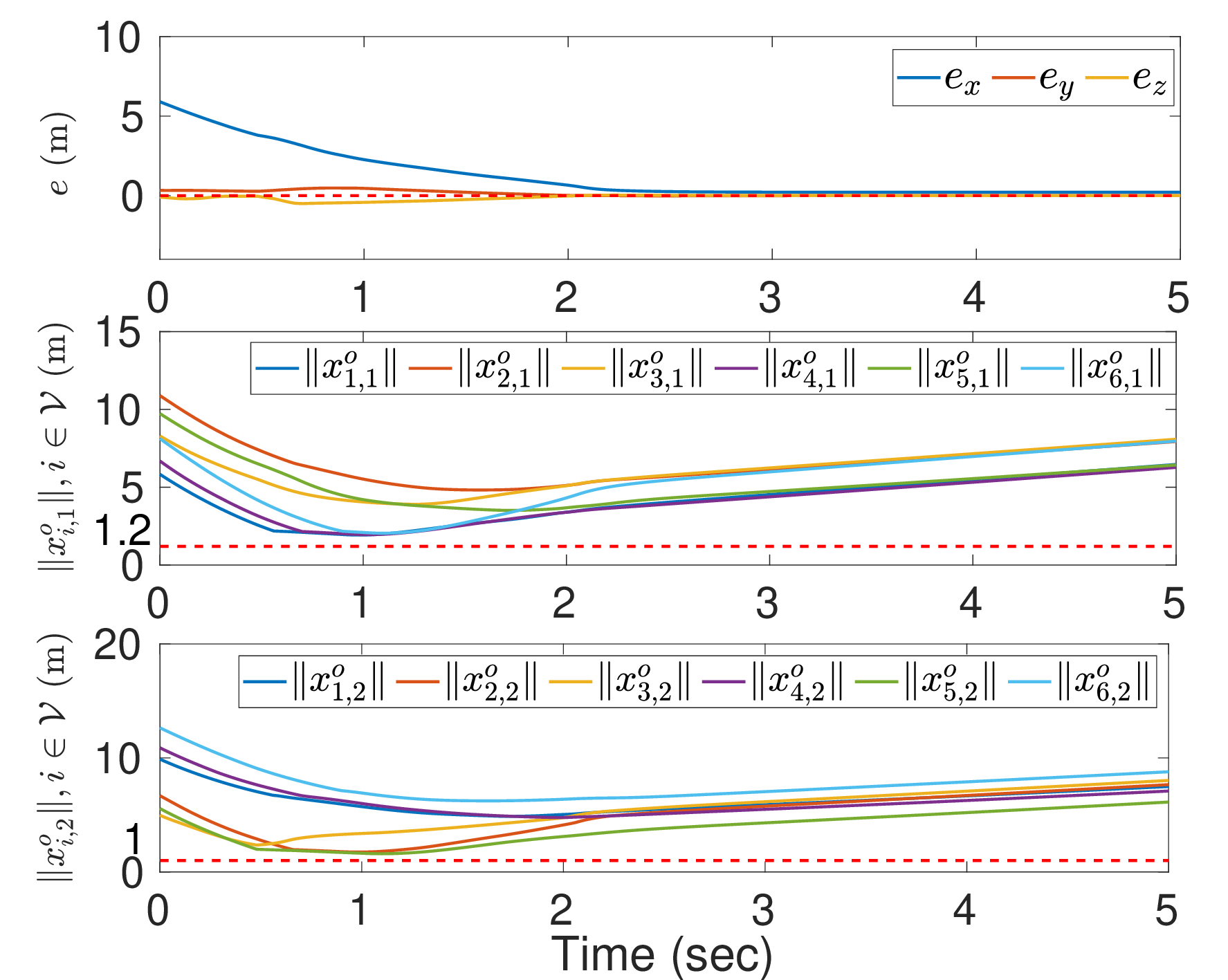}
\caption{ Temporal evolution of the convoying errors $e=[e_{x}, e_y, e_z]\t$ $\in\mathbb{R}^3$, and the robot-obstacles distances $x_{i,1}^o, x_{i,2}^o, i\in\mathcal V,$ in Fig.~\ref{Broken_3D_circle_trajectory} (b) for example.
}
\label{obstacle_dis}
\end{figure}

Additionally, we take Fig.~\ref{3D_line_trajectory} (b) as an example to analyze the state evolution of the {\it ordering-flexible} multi-robot convoying in the 3D Euclidean space. As shown in Fig.~\ref{3D_distance}, the convoying errors $e=[e_x, e_y, e_z]\t\in\mathbb{R}^3$ in three axises converge to zeros, which indicate that the target is eventually convoyed by six robots, i.e., Objective 1) in Definition~\ref{definition_convoying1}. The robot-target distances $\|x_{i,d}\|, i\in\mathcal V,$ and the inter-robot distance $\|x_{i,j}\|, i\neq j\in\mathcal V$ in Fig.~\ref{3D_distance} satisfy $\|x_{i,j}(t)\|\ge 1.5, \|x_{i,d}(t)\|>0, \forall i\neq j \in\mathcal V, t>0$, which thus avoids the overlapping behavior in Objective 3) of Definition~\ref{definition_convoying1}. It is observed in Fig.~\ref{3D_ordering_distance} that the relative distance between the adjacent robots along the ordering sequence $\{3, 5, 1, 6, 4, 2\}$ satisfies $1.5<\lim_{t\rightarrow\infty}\|x_{s[i],s[i+1]}(t)\|<4, s[1]-s[6]:=\{3, 5, 1, 6, 4, 2\}$, which ensures Objective 2) of Definition~\ref{definition_convoying1}. Moreover, Fig.~\ref{3D_inputs} depicts that $u_i=[u_{i,x}, u_{i,y}, u_{i,z}]\t\in\mathbb{R}^3, i\in\mathcal V,$ of robot $i$ are bounded by the limits $\zeta=6$ all the time, i.e., $\|u_i(t)\|_{\infty}\leq 6, \forall t>0$, which guarantee the feasibility of the long-term target convoying  \eqref{constrain_convoying_problem}.

To show the robustness of {\it ordering-flexible} target convoying, Fig.~\ref{Broken_3D_circle_trajectory} illustrates arbitrary two robots suddenly breakdown at $t=0.5 s$ in the six-robot target convoying task. More precisely, it is observed in Fig.~\ref{Broken_3D_circle_trajectory} (a) that the rest four robots $i=1, 2, 3, 5$ still form a successful tetrahedron-pattern convoying with the ordering $\{5,3,1,2\}$ when robots $i=4, 6$ break down. Similarly, the robots $i=1, 3, 4, 6$ achieve the desired tetrahedron-pattern convoying with the ordering $\{3,1,4,6\}$ when robots $i=2, 5$ break down in Fig.~\ref{Broken_3D_circle_trajectory}~(b). It thus verifies  
the robustness of the proposed LTTE algorithm in 3D, which cannot be coped with the previous works~\cite{chen2010surrounding,hu2021bearing,shi2015cooperative,hu2020multiple,sakurama2020multi,liu2019collective,wang2017limit,hu2021distributed2,kou2021cooperative1,kou2021cooperative2,hu2022cooperative}.

Moreover, to verify the ability of obstacle avoidance, we consider two static ball obstacles in the six-robot system according to Remark~\ref{re_obstacle}.
The positions and radiuses of these obstacles are set to be $x_1^o=[2, 8, 0\t]m, r_1^o=1.2m, x_2^o=[2, 13, 0]\t m,  r_2^o=1m$, which does not contain any robots inside at the beginning. Fig.~\ref{Obs_3D_line_trajectory} describes the trajectories of six robots from same initial positions in Fig.~\ref{3D_line_trajectory}~(b) (blue triangles) to obstacle avoiding (yellow triangles), and to the final hexahedron-pattern convoying (red triangles).
It is observed in Fig.~\ref{Obs_3D_line_trajectory} (a) that the dashed trajectories of the six robots successfully deviate from the obstacles (dark blue balls), which implies that the obstacle avoidance are satisfied, whereas Fig.~\ref{Obs_3D_line_trajectory} (b) gives a better illustration of the desired hexahedron-pattern convoying from another view. As shown in Fig.~\ref{obstacle_dis}, the convoying errors $e=[e_x, e_y, e_z]\t\in\mathbb{R}^3$ converge to zeros as well, and the 
the robot-obstacles distances satisfy $x_{i,1}^o(t)>1.2, x_{i,2}^o(t)>1, i\in\mathcal V, \forall t>0$, which verify the obstacle-avoidance capacity in Remark~\ref{re_obstacle}.

\section{Conclusion}
This paper has introduced a LTTE algorithm that achieves the {\it ordering-flexible} multi-robot target convoying by designing and encoding the convoying subtasks as constraints in an online constrained-based optimization framework. Using these constraints, we can perform the long-term target-convoying task under changing environments. The global convergence of the LTTE is analyzed subject to time-varying neighboring collision avoidance and external exponentially vanishing estimation disturbances. The effectiveness, multi-dimensional adaptability and robustness of the proposed LTTE approach are showcased
through 2D {\it ordering-flexible} convoying experiments of three AMRs and 3D simulations.

%\appendix
\section{ Appendix}
\subsection{Appendix A}
\label{Proof_lemma_1}
{\it Proof of Lemma~\ref{lemma_undesired_behavior}.} Firstly, let $m_i-1\in\mathbb{Z}^+$ be the number of neighbor set $\mathcal N_i$ of~robot~$i$, we rewrite the variables and functions in \eqref{constrain_convoying_problem} to be
\begin{align}
\label{vector_variable}
&\delta_i= \big[\delta_{i,0}, \overbrace{0,\dots, 0}^{m_i-1}\big]\t\in\mathbb{R}^{m_i},~ \Phi_i=\big[\phi_{i,0}, \overbrace{
 \dots, \phi_{i,j}}^{m_i-1}\big]\t\in\mathbb{R}^{m_i},\nonumber\\
& \gamma(\Phi_i)=\big[\gamma_1(\phi_{i,0}), \overbrace{\dots, \gamma_2(\phi_{i,j})}^{m_i-1}\big]\t\in\mathbb{R}^{m_i},
\end{align}
one has that the decentralized constraint-based optimization problem \eqref{constrain_convoying_problem} becomes, 
\begin{align}
\label{constrain_convoying_problem1}
&\min\limits_{u_i, \delta_i}\|u_i-\widehat{v}_d^i\|^2+l\|\delta_i\|^2~~\mbox{s.t.}~g_i\leq\mathbf{0}_{m_i}, \mu_i\leq 0,
\end{align}
where $\mathbf{0}_{m_i}=[0,\dots, 0]\t\in\mathbb{R}^{m_i}$, and $g_i, \mu_i$ are given below
\begin{align*}
%\label{contrain_g}
g_i:=\bigg[\frac{\partial \Phi_{i}}{\partial x_i\t} (u_i-\widehat{v}_d^i) + \gamma(\Phi_{i})- \delta_{i} \bigg],~\mu_i:=\|u_i\|_{\infty}-\zeta
\end{align*}
with $\zeta$ being the input limit of robots in \eqref{kinetic_F}. Firstly, since $\phi_{i,j}=r-\|x_{i,j}(0)\|\leq0, \forall i\neq j\in\mathcal V$ in Assumption~\ref{assump_distance}, it follows from Eq.~\eqref{ZCBF_2} that $x_{i}(0)\in\mathcal{T}_{i,j}$.
Using Assumption~\ref{assumption_local_Lipschiz}, it follows from the forward-invariance property in Eq.~\eqref{forward_invariance} that all the states $x_{i}(t)$ will stay in the set $\mathcal{T}_{i,j}$ all along, i.e., 
$x_i(t)\in\mathcal T_{i,j}, \forall i\in\mathcal V, t>0$, which implies that $\|x_{i,j}(t)\|\ge r, \forall i\neq j, t>0$. The inter-robot collision avoidance is thus proved.

Next, we will prevent the robot-target overlapping, i.e., $\|x_{i,d}(t)\|>0, \forall t>0$. Let the corresponding Lagrangian function for robot $i$ in Eq.~\eqref{constrain_convoying_problem1} be $\mathcal L(u_i, \delta_i, \lambda_i, \varsigma_i)=\|u_i-\widehat{v}_d^i\|^2+l\|\delta_i\|^2+\lambda_i\t g_i+\varsigma_i \mu_i$
%\begin{align}
%\label{Lagrangian}
%\mathcal L(u_i, \delta_i, \lambda_i, \varsigma_i)=\|u_i-\widehat{v}_d^i\|^2+l\|\delta_i\|^2+\lambda_i\t g_i+\varsigma_i \mu_i
%\end{align}
with $\lambda_i=[\lambda_{i,1}, \dots, \lambda_{i,m_i}]\t\in\mathbb{R}^{m_i}$ and $\varsigma_i\in\mathbb{R}$.
Using the KKT conditions \cite{boyd2004convex}, we take the partial derivative of $\mathcal L(u_i, \delta_i, \lambda_i, \varsigma_i)$ along the vector $u_i\t, \delta_i\t$,
\begin{align}
\label{KKT}
&\begin{bmatrix}
2(u_i-\widehat{v}_d^i)\\
2l\delta_i
\end{bmatrix}+
\begin{bmatrix}
\frac{\partial \Phi_{i}\t}{\partial x_i} \lambda_{i}+\varsigma_i \mbox{sgn}(\|u_i\|_{\infty}) \\
-\lambda_i
\end{bmatrix}=\mathbf{0}_{n+m_i},\nonumber\\
&\lambda_i\t g_i=0,~\varsigma_i\mu_i=0,~\lambda_{i,k}\geq0, \forall k\in\mathbb{Z}_1^{m_i},
\end{align}
which implies that
\begin{align}
\label{condition_1}
u_i=-l\frac{\partial \Phi_{i}\t}{\partial x_i}\delta_{i}+\widehat{v}_d^i+\frac{\varsigma_i \mbox{sgn}(\|u_i\|_{\infty})}{2},~\lambda_{i}=2l\delta_{i}.
\end{align}
From $\lambda_i\t g_i=0, \varsigma_i\mu_i=0$ in Eq.~\eqref{KKT}, one has that $\lambda_i=\mathbf{0}_{m_i}$ or $g_i=\mathbf{0}_{m_i}$, and $ \varsigma_i=0$ or $\mu_i=0$. 

Since the constraints in \eqref{constrain_convoying_problem1} denote target-approaching and collision-free subtasks which are eventually activated during the process, then the condition of $\lambda_i=\mathbf{0}_{m_i}$ can be excluded after a constant time $T_1>0$, it implies that $g_i(t)=\mathbf{0}_{m_i}$ when $t>T_1$. Moreover, the control-input constraint is activated (i.e., $\mu_i=0$) when the robots are far away from the target, which implies that there exists a time $T_2>0$ such that $\varsigma_i=0$ holds when $t>T_2$.
Let $T_3:=\max\{T_1, T_2\}$,  one has that $ \|x_{i,j}(t)\|\ge r$, $\|x_{i,d}(t)\|>0, \forall i\neq j \in\mathcal V, t\in[0, T_3]$ because $u_i$ is limited and the constraints $g_i$ are not activated. Next, we will prove the undesirable overlapping when $t>T_3$. For $g_i=\mathbf{0}_{m_i}$ and $\varsigma_i=0$, it follows from~\eqref{condition_1} that 
\begin{align}
\label{condition_2}
-l\frac{\partial \Phi_{i}}{\partial x_i\t} \frac{\partial \Phi_{i}\t}{\partial x_i}\delta_{i} + \gamma(\Phi_{i})- \delta_{i} =\mathbf{0}_{m_i}.
\end{align}
From the fact  ${\partial \Phi_{i}}/{\partial x_i\t}=({\partial \Phi_{i}\t}/{\partial x_i})\t \in\mathbb{R}^{m\times n}$, it follows from that \eqref{condition_2} that 
\begin{align}
\label{condition_2_1}
 \delta_{i}=\Xi_i  \gamma(\Phi_{i})
\end{align}
with 
\begin{align}
\label{inverse_matrix}
\Xi_i=\bigg(I_m+l\frac{\partial \Phi_{i}}{\partial x_i\t} \Big(\frac{\partial \Phi_{i}}{\partial x_i\t}\Big)\t \bigg)^{-1}\in\mathbb{R}^{m\times m}.
\end{align}
Substituting $\varsigma_i=0$ and Eq.~\eqref{condition_2_1} into Eq.~\eqref{condition_1} yields the optimal input
\begin{align}
\label{opt_input}
u_i^*=-l\frac{\partial \Phi_{i}\t}{\partial x_i}\Xi_i  \gamma(\Phi_{i})+\widehat{v}_d^i.
\end{align}
Based on the implicit function $\gamma(\Phi_i), i\in\mathcal V$ in Eq.~\eqref{vector_variable}, we pick a Lyapunov function candidate below
\begin{align}
\label{V_fun}
V=\sum_{i=1}^N \gamma_1(\phi_{i,0})^2+\frac{1}{2}\sum_{i=1}^N\sum_{j\in\mathcal N_i}
\gamma_2(\phi_{i,j})^2,
\end{align}
whose derivative is 
\begin{align}
\label{dot_V_fun}
\dot{V}=&\sum_{i=1}^N\gamma_1(\phi_{i,0})\frac{\partial \gamma_1(\phi_{i,0})}{\partial x_{i,d}\t}(\dot{x}_i-\dot{x}_d)\nonumber\\
&+\frac{1}{2}\sum_{i=1}^N\sum_{j\in\mathcal N_i}\gamma_2(\phi_{i,j})\frac{\partial \gamma_2(\phi_{i,j})}{\partial x_{i,j}\t}(\dot{x}_i-\dot{x}_j).
\end{align}
Recalling the definition of $\phi_{i,j}$ in Eq.~\eqref{convoying_condition2}, one has that
%\begin{align*}
%%\label{in_equality1}
%&\gamma_2(\phi_{i,j})\frac{\partial \gamma_2(\phi_{i,j})}{\partial x_{i,j}\t}\dot{x}_i=-\gamma_2(\phi_{i,j})\frac{\partial \gamma_2(\phi_{i,j})}{\partial x_{i,j}\t}\dot{x}_j, \forall j\neq i,
%\end{align*}
%which implies that
\begin{align}
\label{in_equality2}
&\sum_{i=1}^N\sum_{j\in\mathcal N_i}\gamma_2(\phi_{i,j})\frac{\partial \gamma_2(\phi_{i,j})}{\partial x_{i,j}\t}(\dot{x}_i-\dot{x}_j)\nonumber\\
=&2\sum_{i=1}^N\sum_{j\in\mathcal N_i}\gamma_2(\phi_{i,j})\frac{\partial \gamma_2(\phi_{i,j})}{\partial x_{i,j}\t}\dot{x}_i.
\end{align}
Moreover, we have $\sum_{i=1}^N\sum_{j\in\mathcal N_i}\gamma_2(\phi_{i,j})({\partial \gamma_2(\phi_{i,j})}/{\partial x_{i,j}\t})$ $v_d=0$.
%\begin{align}
%\label{in_equality3}
%\sum_{i=1}^N\sum_{j\in\mathcal N_i}\gamma_2(\phi_{i,j})\frac{\partial \gamma_2(\phi_{i,j})}{\partial x_{i,j}\t}v_d=0.
%\end{align}
Substituting Eqs.~\eqref{kinetic_F}, \eqref{kinetic_target}, \eqref{in_equality2} into Eq.~\eqref{dot_V_fun} yields 
\begin{align}
\label{dot_V_fun1}
\dot{V}=&\sum_{i=1}^N\bigg\{\bigg(\gamma_1(\phi_{i,0})\frac{\partial \gamma_1(\phi_{i,0})}{\partial x_{i,d}\t}\nonumber\\
             &+\sum_{j\in\mathcal N_i}\gamma_2(\phi_{i,j})\frac{\partial \gamma_2(\phi_{i,j})}{\partial x_{i,j}\t}\bigg)(u_i-v_d)\bigg\}.
\end{align}
Meanwhile, from the definition of $\phi_{i,0}, \phi_{i,j}$ in Eqs.~\eqref{ZCBF_1} and \eqref{ZCBF_2}, one has that
\begin{align}
\label{partial_transtion}
\frac{\partial \gamma_1(\phi_{i,0})}{\partial x_{i,d}\t}=\frac{\partial \gamma_1(\phi_{i,0})}{\partial \phi_{i,0}} \frac{\partial \phi_{i,0}}{\partial x_{i}\t}, \frac{\partial \gamma_2(\phi_{i,j})}{\partial x_{i,j}\t}=\frac{\partial \gamma_2(\phi_{i,j})}{\partial \phi_{i,j}} \frac{\partial \phi_{i,j}}{\partial x_i\t}.
\end{align}
Recalling $\gamma_1(\cdot), \gamma_{2}(\cdot)$ in Eqs.~\eqref{gamma_condition1} and \eqref{gamma_condition2} are locally Lpischiz, there exists a constant $B>0$ satisfying 
\begin{align}
\label{bound_gamma}
\frac{\partial \gamma_1(\phi_{i,0})}{\partial \phi_{i,0}\t}\leq B, \frac{\partial \gamma_2(\phi_{i,j})}{\partial \phi_{i,j}\t}\leq B, \forall i\in\mathcal V, j \in\mathcal N_i,
\end{align}
it then follows from Eqs.~\eqref{vector_variable},~\eqref{opt_input},~\eqref{partial_transtion} and \eqref{bound_gamma} that $\dot{V}$ in Eq.~\eqref{dot_V_fun1} rewrites 
\begin{align}
\label{dot_V_fun2}
%\leq&B\sum_{i=1}^N\gamma(\Phi_i)\t\frac{\partial \Phi_i}{\partial x_i\t}\bigg(-l\frac{\partial \Phi_{i}\t}{\partial x_i}\Xi_i  \gamma(\Phi_{i})+\widetilde{v}_d^i\bigg)\nonumber\\
\dot{V}	   \leq&-B\sum_{i=1}^N\bigg(\frac{\partial \Phi_i\t}{\partial x_i} \gamma(\Phi_i)\bigg)\t \bigg(l \frac{\partial \Phi_{i}\t}{\partial x_i} \Xi_i  \gamma(\Phi_{i})-\widetilde{v}_d^i\bigg)
\end{align}
with $\widetilde{v}_d^i:=\widehat{v}_d^i-v_d$. 
Meanwhile, according to Woodbury matrix identity \cite{higham2002accuracy}, one has that $\Xi_i$ in \eqref{inverse_matrix} is rewritten as 
\begin{align}
\label{transver_xi}
\Xi_i=I_{m_i}-l\frac{\partial \Phi_{i}}{\partial x_i\t} \widetilde{\Xi}_i\frac{\partial \Phi_{i}}{\partial x_i\t}
\end{align}
with
\begin{align}
\label{inverse_xi}
\widetilde{\Xi}_i:=\bigg(I_n+l\Big(\frac{\partial \Phi_{i}}{\partial x_i\t}\Big)\t\frac{\partial \Phi_{i}}{\partial x_i\t} \bigg)^{-1}\in\mathbb{R}^{n\times n}.
\end{align}
Combining Eqs.~\eqref{transver_xi} and \eqref{inverse_xi} together, one has that  $({\partial \Phi_{i}\t}/{\partial x_i}) \Xi_i = \widetilde{\Xi}_i ({\partial \Phi_{i}\t}/{\partial x_i})$.
Since the matrix $\widetilde{\Xi}_i$ in Eq.~\eqref{inverse_matrix} is positive definite, one has that the smallest eigenvalue of $\widetilde{\Xi}_i$ satisfies $\lambda_{min}(\widetilde{\Xi}_i)>0$, which implies that  
\begin{align}
\label{inequality_1}
\bigg(\frac{\partial \Phi_i\t}{\partial x_i} \gamma(\Phi_i)\bigg)\t \widetilde{\Xi}_i \bigg(\frac{\partial \Phi_{i}\t}{\partial x_i} \gamma(\Phi_{i})\bigg)\geq \lambda_{min}(\widetilde{\Xi}_i)\bigg\|\frac{\partial \Phi_i\t}{\partial x_i} \gamma(\Phi_{i})\bigg\|^2
\end{align}
with Courant-Fischer Theorem~\cite{parlett1998symmetric}. Meanwhile, one has
\begin{align}
\label{inequality_2}
\bigg(\frac{\partial \Phi_i\t}{\partial x_i} \gamma(\Phi_{i})\bigg)\t \widetilde{v}_d^i \leq \frac{1}{\kappa}\bigg\|\frac{\partial \Phi_i\t}{\partial x_i} \gamma(\Phi_{i})\bigg\|^2 + \kappa\|\widetilde{v}_d^i\|^2
\end{align}
with an arbitrarily selected constant $\kappa>0$ satisfying ${1}/{\kappa}< l\lambda_{min}(\widetilde{\Xi}_i)$, which then follows from Eqs.~\eqref{dot_V_fun2},~\eqref{inequality_1},~\eqref{inequality_2} that $\dot{V}\leq-\alpha + \beta$
with
\begin{align}
\label{value_1}
\alpha:=&B\sum\limits_{i\in\mathcal V}\bigg(l\lambda_{min}(\widetilde{\Xi}_i)-\frac{1}{\kappa}\bigg)\bigg\|\frac{\partial \Phi_i\t}{\partial x_i} \gamma(\Phi_i)\bigg\|^2\geq0,\nonumber\\
\beta:=&B\sum_{i\in\mathcal V} \kappa\|\widetilde{v}_d^i\|^2\geq 0.
\end{align}
Using the comparison theorems \cite{khalil2002nonlinear} and integrating $\dot{V}$ in Eq.~\eqref{dot_V_fun2} along the time $t$ yields 
\begin{align}
\label{integra_V}
V(t)\leq V(T_3)-\int_{T_3}^t \alpha(s)ds + \int_{T_3}^t\beta(s)ds,~t>T_3.
\end{align}
Using Assumption~\ref{assump_distance}, it follows from Eqs.~\eqref{gamma_condition1}, \eqref{gamma_condition2} and \eqref{V_fun} that the initial value $V(T_3)$ is bounded.
From Eq.~\eqref{value_1}, one has that $-\int_{T_3}^t \alpha(s)ds\leq0$.
Recalling $\lim_{t\rightarrow\infty} \widetilde{v}_d^i(t)=\mathbf{0}_n,$ exponentially in Remark~\ref{target_vel_convergence}, 
it follows from Eq.~\eqref{value_1} that $\lim_{t\rightarrow\infty}\beta(t)=0$ exponentially, which implies that $\int_{T_3}^t\beta(s)ds$ is upper bounded. Then, $V(t)$ is bounded.

Since $V(t)\rightarrow\infty$ only if there exists $\|x_{i,j}(t)\|=0, \forall j\in\mathcal N_i$ or $\|x_{i,d}(t)\|>0$, it contradicts with the bounded $V(t)$ in \eqref{integra_V}, which implies that $ \|x_{i,j}(t)\|\geq r, \|x_{i,d}(t)\|>0, \forall i\neq j \in\mathcal V, \forall t>0$. The proof is thus completed.
\eproof

\subsection{Appendix B}
\label{Proof_lemma_2}
{\it Proof of Lemma~\ref{lemma_convoying}.} Since $V(t)$ is upper bounded in Lemma~\ref{lemma_undesired_behavior}, it follows from Eqs.~\eqref{vector_variable} and \eqref{V_fun} that $\gamma(\Phi_i), \Phi_i$ are all bounded.
Moreover, from the definition of $\alpha$ in Eq.~\eqref{value_1}, $V(t)$ in Eq.~\eqref{integra_V} can be rewritten below,
\begin{align}
\label{integra_V1}
\int_{T_3}^t \alpha(s)ds\leq V(T_3)-V(t)+\int_{T_3}^t\beta(s)ds,
\end{align}
which implies that $\int_{T_3}^t \alpha(s)ds$ is upper bounded and has a finite value when $t\rightarrow\infty$. Since $\dot{\alpha}$ contains ${ \gamma}(\Phi_i), {\dot{ \gamma}(\Phi_i)}, \dot{\Phi}_i, \ddot{\Phi}_i$ which are bounded, one has that $\alpha$ is uniformly continous. It then follows from Barbalat's lemma~\cite{khalil2002nonlinear} that $\lim_{t\rightarrow\infty}\alpha(t)=0$.
%\begin{align*}
%%\label{limit_alpha}
%\lim_{t\rightarrow\infty}\alpha(t)=0.
%\end{align*}
From the definition of~$\alpha$ in Eq.~\eqref{value_1}, one has that
\begin{align}
\label{lemma1_converge1}
\lim_{t\rightarrow\infty}\frac{\partial \Phi_i\t(t)}{\partial x_i(t)} \gamma(\Phi_{i}(t))=\mathbf{0}_{n}, \forall i\in\mathcal V. 
\end{align}
It, together with $\Phi_i, \gamma(\Phi_i)$ defined in Eq.~\eqref{vector_variable}, gives 
\begin{align}
\label{lemma1_converg2}
\lim_{t\rightarrow\infty} &\bigg\{\frac{\partial \phi_{i,0}(t)}{\partial x_i\t(t)}\gamma_1(\phi_{i,0}(t))+\sum_{j\in\mathcal N_i}\frac{\partial \phi_{i,j}(t)}{\partial x_i\t(t)} \gamma_2(\phi_{i,j}(t))\bigg\}=\mathbf{0}_{n}. 
\end{align}
According to the definition of $\phi_{i,0}=-h_{i.0}, \phi_{i,j}=-h_{i,j}, j\in\mathcal N_i$ in Eqs.~\eqref{ZCBF_1} \eqref{ZCBF_2}, one has that 
\begin{align}
\label{partial_phi}
\frac{\partial \phi_{i,0}(t)}{\partial x_i\t(t)}=\frac{x_{i,d}\t}{\|x_{i,d}\|},~\frac{\partial \phi_{i,j}(t)}{\partial x_i\t(t)}=\frac{-x_{i,j}}{\|x_{i,j}\|}, j\in\mathcal N_i.
\end{align}
It follows from Eqs.~\eqref{lemma1_converg2} and \eqref{partial_phi} that
\begin{align}
\label{lemma1_converg3}
\lim_{t\rightarrow\infty}&\Big\{\frac{x_{i,d}(t)}{\|x_{i,d}(t)\|}\gamma_1\big(\|x_{i,d}(t)\|\big)-\sum_{j\in\mathcal N_i}\frac{x_{i,j}(t)}{\|x_{i,j}(t)\|}\gamma_2\big(r\nonumber\\
&-\|x_{i,j}(t)\|\big)\Big\}=\mathbf{0}_n, i\in\mathcal V.
\end{align}
For arbitrary two neighboring robots $i\neq j\in\mathcal V$, one has $\gamma_2\big(r-\|x_{i,j}\|\big){x_{i,j}}/{\|x_{i,j}\|}=-\gamma_2\big(r-\|x_{j,i}\|\big){x_{j,i}}/{\|x_{j,i}\|}$,
%\begin{align*}
%%\label{neighboring_condition}
%\frac{x_{i,j}}{\|x_{i,j}\|}\gamma_2\big(r-\|x_{i,j}\|\big)=-\frac{x_{j,i}}{\|x_{j,i}\|}\gamma_2\big(r-\|x_{j,i}\|\big),
%\end{align*}
which implies that $\sum_{i=1}^N\sum_{j\in\mathcal N_i}\gamma_2\big(r-\|x_{i,j}(t)\|\big){x_{i,j}(t)}$ $/{\|x_{i,j}(t)\|}=\mathbf{0}_n, \forall t>0.$
%\begin{align}
%\label{sum_neighboring_condition}
%\sum_{i=1}^N\sum_{j\in\mathcal N_i}\frac{x_{i,j}(t)}{\|x_{i,j}(t)\|}\gamma_2\big(r-\|x_{i,j}(t)\|\big)=\mathbf{0}_n, \forall t>0.
%\end{align}
Then, it follows from Eqs.~\eqref{lemma1_converg2} and \eqref{lemma1_converg3} that $\lim_{t\rightarrow\infty}\sum_{i=1}^N \{\gamma_1\big(\|x_{i,d}(t)\|\big) {x_{i,d}}/{\|x_{i,d}\|}\}=\mathbf{0}_n.$
%\begin{align*}
%\lim_{t\rightarrow\infty}\sum_{i=1}^N \frac{x_{i,d}}{\|x_{i,d}\|}\gamma_1\big(\|x_{i,d}(t)\|\big)=\mathbf{0}_n.
%\end{align*} 

Since $\gamma_1(\|x_{i,d}(t)\|)$ in Eq.~\eqref{gamma_condition1} is an odd function, one has that $\lim_{t\rightarrow\infty}\sum_{i=1}^N x_{i,d}(t)=\mathbf{0}_n.$
%\begin{align}
%\label{converge_convoy}
%\lim_{t\rightarrow\infty}\sum_{i=1}^N x_{i,d}(t)=\mathbf{0}_n.
%\end{align}
\begin{figure}[!htb]
\centering
\includegraphics[width=7.5cm]{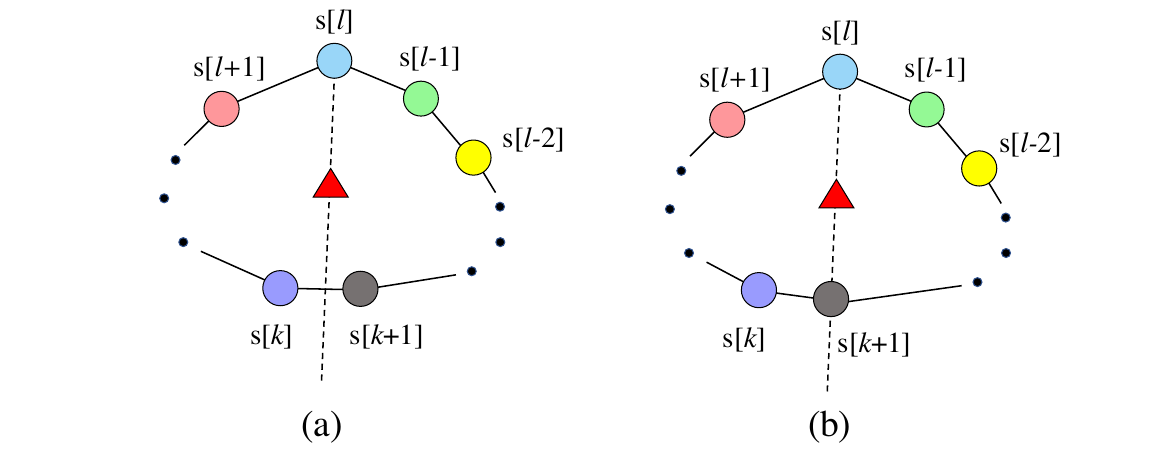}
\caption{ 2D illustration of two-subgroup robots along the dashed line formed by $x_s[l]$ and $x_d$.
}
\label{ordering_flexible_proof}
\end{figure}
Let the target-convoying errors be $e:=1/N\sum_{i=1}^Nx_i-x_d$, one has that 
%$\lim_{t\rightarrow\infty}e(t)%=&\lim_{t\rightarrow\infty}\Big\{ \frac{1}{N} \sum_{i=1}^Nx_{i}(t)-x_d(t)\Big\}\nonumber\\
%=\lim_{t\rightarrow\infty} {1}/{N} \sum_{i=1}^N\{x_{i}(t)-x_d(t)\}=\mathbf{0}_n$
\begin{align}
\label{converge_convoy_err}
\lim_{t\rightarrow\infty}e(t)%=&\lim_{t\rightarrow\infty}\Big\{ \frac{1}{N} \sum_{i=1}^Nx_{i}(t)-x_d(t)\Big\}\nonumber\\
=&\lim_{t\rightarrow\infty} \frac{1}{N} \sum_{i=1}^N\Big\{x_{i}(t)-x_d(t)\Big\}=\mathbf{0}_n
\end{align}
with $1/N\sum_{i=1}^N x_d=x_d$.
The proof of convex-hull convoying is thus completed.
\eproof

\subsection{ Appendix C}
\label{Proof_lemma_3}
{\it Proof of Lemma~\ref{lemma_ordering_pattern}.} From Eqs.~\eqref{opt_input} and~\eqref{lemma1_converge1} in the proof of Lemmas~\ref{lemma_undesired_behavior} and \ref{lemma_convoying}, one has that $\lim_{t\rightarrow\infty}u_i^*(t)=\widehat{v}_d^i, \forall i\in\mathcal V.$
%\begin{align*}
%\lim_{t\rightarrow\infty}u_i^*(t)=\widehat{v}_d^i, \forall i\in\mathcal V.
%\end{align*}
Combining with the fact $\lim_{t\rightarrow\infty}{v}_d^i(t)=v_d(t)$ in Remark \ref{target_vel_convergence}, one has that the optimal input $u_i^*$ of robot $i$ satisfy $\lim_{t\rightarrow\infty}u_i^*(t)=\lim_{t\rightarrow\infty}u_j^*(t)=v_d(t), \forall i\neq j\in\mathcal V,$
%\begin{align*}
%\lim_{t\rightarrow\infty}u_i^*(t)=\lim_{t\rightarrow\infty}u_j^*(t)=v_d(t), \forall i\neq j\in\mathcal V,
%\end{align*}
which follows from Eq.~\eqref{kinetic_F} that $\lim_{t\rightarrow\infty}\{\dot{x}_i(t)-\dot{x}_j(t)\}=\mathbf{0}_n$, (i.e., the relative position of $x_i$ against any other $x_j, j\neq i$ converges to be time-invariant and a rigid pattern with a spatial sequence $ s[1], s[2], \cdots, s[N]$ is formed). The condition (a) in \eqref{condition_pattern} is thus verified.

Next, we will prove the condition (b) in Eq.~\eqref{condition_pattern}. From Lemma~\ref{lemma_undesired_behavior}, one has that  $\|x_{i,j}(t)\|\ge r, \|x_{i,d}(t)\|>0, \forall i\neq j \in\mathcal V, t>0$, which implies that
\begin{align}
&\lim_{t\rightarrow\infty}\|x_{s[i], s[i+1]}(t)\| \geq r,\nonumber\\
&\forall i\in\mathbb{Z}_1^{N}~(\mbox{If}~i=N, \mbox{then}~s[i+1]=s[1]).
\end{align}
Then, the left-side inequality in Eq.~\eqref{condition_pattern} is fulfilled. To proceed, the right-side inequality of condition (b) in Eq.~\eqref{condition_pattern} is proved by contradiction.

Intuitively, from Lemma~\ref{lemma_convoying}, we obtain that the target $x_d$ is asymptotically convoyed in the centroid of the convex hull $\sum_{i=1}^N x_i/N$ with a spatial sequence $ s[1], s[2], \cdots, s[N]$. Without loss of generality, we assume that there exists at least one pair of adjacent robots labelled $s[l],s[l+1], \forall l\in\mathbb{Z}_1^{N-1}$ such that $\|x_{s[l],s[l+1]}\|> R$, and then separate the robots $\mathcal V$ into two subgroups $\mathcal V_1:=\{s[l+1], s[l+2], \cdots, s[k]\}$ and $\mathcal V_2:=\{s[l-1], s[l-2], \cdots, s[k+1]\}$ along the dashed line (i.e., a dashed plane in 3D) formed by $x_{s[l]}$ and $x_d$, as shown in Fig.~\ref{ordering_flexible_proof}.
Then, the contradiction is analyzed in the following two cases.

%It, together with the fact $\|x_{i,d}(t)\|\neq0$ in Lemma~\ref{lemma_undesired_behavior}, gives that all robots finally form a convex hull $\sum_{i\in\mathcal V}c_ix_i(t)$. 

Case 1: \{Robot $s[k+1]$ is not on the dashed line, see Fig.~\ref{ordering_flexible_proof}~(a)\}. For $\|x_{s[l],s[l+1]}\|> R$, one has that $\|x_{s[l],j}\|> R, j\in\mathcal V_1$, which implies that robot $s[l]$ may only have neighbors fulfilling 
$\|x_{s[l], o}\|\leq R, o\in\mathcal V_2$. Moreover, the vectors $x_{s[l],d}$ and $x_{s[l], o}, o\in\mathcal V_2,$ are not in same or opposite direction, i.e., 
\begin{align}
\label{No_same_oppo_dir}
-1<\lim_{t\rightarrow\infty}\frac{x_{s[l],d}(t)\cdot x_{s[l], o}(t)}{\|x_{s[l],d}(t)\|\|x_{s[l], o}(t)\|}<1,  o\in\mathcal V_2,
\end{align}
 Let $\vartheta_o, o\in\mathcal V_2,$ be the repulsion term between robots $s[l]$ and $s[o], o\in\mathcal V_2$ below, 
\begin{align}
\label{vartheta_repul}
\vartheta_o=\frac{ x_{s[l], o}}{\|x_{s[l], o}\|}\gamma_2\big(r-\|x_{s[l],o}(t)\|\big), o\in\mathcal V_2,
\end{align}
it follows from Eq.~\eqref{No_same_oppo_dir} that there there only exist left-side repulsion terms $\vartheta_o^L, o\in\mathcal V_2,$ perpendicular to $x_{s[l], d}$ below, 
\begin{align}
\label{left_repul_item}
\vartheta_o^L=\vartheta_o-\frac{x_{s[l],d}}{\|x_{s[l],d}\|} \cdot \vartheta_o.
\end{align}
However, $\vartheta_o^L$ cannot be eliminated by the robot-target attraction term $({x_{s[l],d}}/{\|x_{s[l],d}\|})\gamma_1\big(\|x_{s[l],d}\|\big)$. Then, one has 
\begin{align*}
\lim_{t\rightarrow\infty}&\bigg\{\frac{x_{s[l],d}(t)}{\|x_{s[l],d}(t)\|}\gamma_1\big(\|x_{s[l],d}(t)\|\big)-\sum_{o\in\mathcal V_2}\vartheta_o\bigg\}\neq\mathbf{0}_n.
\end{align*}
Then, it contradicts the condition in~Eq.~\eqref{lemma1_converg3}.

Case 2: \{Robot $s[k+1]$ is on the dashed line, see Fig.~\ref{ordering_flexible_proof}~(b)\}.
Since robot $s[k+1]$ is on the dashed line, one has that $x_{s[l], s[k+1]}$ and $x_{s[l],d}$ are in the opposite direction, which follows from Eqs.~\eqref{vartheta_repul} and \eqref{left_repul_item} that 
$\vartheta_{s[k+1]}^L=\vartheta_{s[k+1]}-({x_{s[l],d}}/{\|x_{s[l],d}\|}) \cdot \vartheta_o=\mathbf{0}_n$. Then, the repulsion term $\vartheta_{s[k+1]}$ cannot eliminate the other left-side repulsion term $\vartheta_o^L$ in Fig.~\ref{ordering_flexible_proof}~(b) as well. The following contradiction analysis is the same as Case 1, which is omitted. Finally, we obtain that $\lim_{t\rightarrow\infty}\|x_{s[i], s[i+1]}(t)\|\leq R, \forall i\in\mathbb{Z}_1^{N}~(\mbox{If}~i=N, \mbox{then}~s[i+1]=s[1]).$
%\begin{align*}
%&\lim_{t\rightarrow\infty}\|x_{s[i], s[i+1]}(t)\|\leq R,\nonumber\\
%&\forall i\in\mathbb{Z}_1^{N}~(\mbox{If}~i=N, \mbox{then}~s[i+1]=s[1]).
%\end{align*}
The proof is thus completed.
\eproof

\bibliographystyle{IEEEtran}
\bibliography{IEEEabrv,ref}

\end{document}